\documentclass[preprint,12pt]{elsarticle}

\usepackage{amssymb}

\usepackage{amsmath}
\biboptions{sort&compress}
\usepackage{hyperref}
\hypersetup{
    colorlinks=true,
    linkcolor=blue,
    urlcolor=blue,
    citecolor=blue}
\hypersetup{pdfauthor={Name}}

\usepackage{algorithm}
\usepackage[rightComments=true]{algpseudocodex}

\usepackage{multirow}

\usepackage{booktabs}
\usepackage{makecell}
\usepackage{subcaption}
\usepackage{caption}
\journal{Applied Soft Computing}

\begin{document}

\begin{frontmatter}



\title{Cooperative Coevolution for Non-Separable Large-Scale Black-Box Optimization: Convergence Analyses and Distributed Accelerations}

\author[label_HIT,label_SUSTech]{Qiqi Duan\fnref{label_equal}}
\author[label_UTS,label_SUSTech]{Chang Shao\fnref{label_equal}}
\author[label_SUSTech]{Guochen Zhou}
\author[label_SUSTech]{Haobin Yang}
\author[label_SUSTech]{Qi Zhao}
\author[label_SUSTech]{Yuhui Shi\corref{cor1}}

\affiliation[label_HIT]{organization={Harbin Institute of Technology},
            city={Harbin},
            postcode={150001},
            state={Heilongjiang},
            country={China}}
\affiliation[label_UTS]{organization={Australian Artificial Intelligence Institute (AAII), Faculty of Engineering and Information Technology, University of Technology Sydney},
            addressline={15 Broadway, Ultimo},
            city={Sydney},
            postcode={2007},
            state={NSW},
            country={Australia}}
\affiliation[label_SUSTech]{organization={Department of Computer Science and Engineering, Southern University of Science and Technology},
            city={Shenzhen},
            postcode={518055},
            state={Guangdong},
            country={China}}
\fntext[label2]{Qiqi Duan and Chang Shao contributed equally.}

\cortext[cor1]{Corresponding author.}
\ead{shiyh@sustech.edu.cn}

\begin{abstract}
Given the ubiquity of non-separable optimization problems in real worlds, in this paper we analyze and extend the large-scale version of the well-known cooperative coevolution (CC), a divide-and-conquer black-box optimization framework, on non-separable functions. First, we reveal empirical reasons of when decomposition-based methods are preferred or not in practice on some non-separable large-scale problems, which have not been clearly pointed out in many previous CC papers. Then, we formalize CC to a continuous-game model via simplification, but without losing its essential property. Different from previous evolutionary game theory for CC, our new model provides a much simpler but useful viewpoint to analyze its convergence, since only the pure Nash equilibrium concept is needed and more general fitness landscapes can be explicitly considered. Based on convergence analyses, we propose a hierarchical decomposition strategy for better generalization, as for any decomposition, there is a risk of getting trapped into a suboptimal Nash equilibrium. Finally, we use powerful distributed computing to accelerate it under the recent multi-level learning framework, which combines the fine-tuning ability from decomposition with the invariance property of CMA-ES. Experiments on a set of high-dimensional test functions validate both its search performance and scalability (w.r.t. CPU cores) on a clustering computing platform with 400 CPU cores.
\end{abstract}



\end{frontmatter}



\section{Introduction}

Recent advances in artificial intelligence (particularly deep models \cite{J_Nature_lecun2015deep,J_NN_schmidhuber2015deep,arXiv_zador2022nextgeneration}) and big data have generated a growing number of large-scale optimization (LSO) problems, including challenging high-dimensional black-box optimization (BBO) instances \cite{J_FoCM_nesterov2017random,J_OMS_hansen2021coco}. An \emph{OpenAI} team \cite{arXiv_salimans2017evolution} used a highly-parallelized version of evolution strategies (ES) to optimize millions of weights of deep neural networks \cite{J_Nature_mnih2015humanlevel} for direct policy search in reinforcement learning (RL). In the recent \emph{Science} paper, Fan et al. \cite{J_Science_fan2020highresolution} hybridized genetic algorithm (GA) and simulated annealing (SA) to solve a big-data-based paleobiology model on a high-performing supercomputer called Tianhe. A \emph{DeepMind} team proposed a population-based training method for hyper-parameter optimization of generative adversarial networks \cite{arXiv_jaderberg2017population}, multi-agent RL \cite{J_Science_jaderberg2019humanlevel}, and a deep vision model for self-driving cars \cite{D_How}.

Although evolutionary algorithms (EAs \cite{J_Science_forrest1993genetic,J_Nature_eiben2015evolutionary,J_NatMI_miikkulainen2021biological}) are one very popular algorithm family for BBO, nearly all standard versions of EAs have to be significantly improved in the LSO context, since they suffer easily from the notorious \emph{curse of dimensionality} (from their random sampling operations). For a survey of EAs for LSO, see e.g., \cite{C_PPSN_varelas2018comparative,J_TEVC_omidvar2022reviewa,J_TEVC_omidvar2022review} and references wherein. For efficient search in large-scale space, exploiting the possibly useful problem structure is a critical step to accelerate convergence progress, as shown in Nesterov’s outstanding optimization book \cite{B_nesterov2018lectures}. In this paper, we focus on mainly two of the representative EAs both with some successful LSO applications: 1) cooperative co-evolution (CC \cite{C_PPSN_potter1994cooperative,B_potter1997design,J_ECJ_potter2000cooperative,J_SMO_gandomi2023variable,J_InfSci_yang2008large,J_TEVC_ma2019survey}) based on the (co-adapted) modularity assumption; and 2) covariance matrix adaptation evolution strategies (CMA-ES \cite{J_ECJ_hansen2003reducing,J_ECJ_akimoto2020diagonal}), which is regarded widely as the state-of-the-art for BBO in a recent \emph{Nature} review for EAs \cite{J_Nature_eiben2015evolutionary}. For modern CC, CMA-ES is often chosen as the basic suboptimizer for all subproblems, because of its well-studied theoretical properties (invariance against affine transformation \cite{J_JMLR_ollivier2017informationgeometric}, unbiases under neutral selection \cite{BS_hansen2014principled}, maximum entropy (diversity) principle \cite{J_NACO_beyer2002evolution}, learning of natural gradient \cite{J_JMLR_wierstra2014natural,J_Algo_akimoto2012theoretical}) and generalizable search abilities in particular on \emph{non-separable} and \emph{ill-conditioned} optimization problems.

In 2014, a modern CC variant with the so-called differential grouping (DG) technique \cite{J_TEVC_omidvar2014cooperative} obtained promising results on a class of \emph{partially additively separable} (PAS) \cite{J_ECJ_muhlenbein1999fdaa} benchmark functions for LSO (mainly from IEEE-CEC competitions \cite{R_tang2009benchmark,R_li2013benchmark,J_InfSci_omidvar2015designing}). Its well-established theoretical foundation to detect variable interactions has now sparked many following-up works (see Section 2.A). However, as previously argued in a \emph{Nature} review \cite{J_Nature_bonabeau2000inspiration}, “\emph{most instances of most problems are not readily ‘linearly’ decomposable into building blocks}”. Indeed, a plenty of real-world problems have complex (\emph{nonlinear}) objective functions, where typically there have \emph{explicit} or \emph{implicit} interactions between any two variables for both gradient-based optimization \cite{J_TMTT_feng2016parallel} and BBO. Take e.g., \cite{J_RSIF_cheney2018scalable,J_TEVC_farahmand2010interaction,C_PPSN_vidal2010threshold,C_GECCO_rainville2013sustainable,J_TPAMI_zhai2016making,J_TEVC_he2016cooperative,J_TNNLS_fan2017collective,J_TNNLS_gong2021evolving,J_TMIS_rashid2022anomaly,J_TEVC_liu2022surrogateassisted,J_TCYB_zhao2020evolutionary} as examples, to name a few\footnote{Here we provide an online website (\url{https://tinyurl.com/3d4nnneb}) which covers many real-world applications (nearly all of them are non-separable). What kinds of real-world applications are PAS is still an open question.}. Even linear regression, perhaps the simplest data model, has a \emph{fully non-separable} loss function form\footnote{Only when the involved data has a block matrix structure, the resulting objective function is PAS. To our knowledge, however, little of real-world applications exhibit this (see \url{https://archive.ics.uci.edu/ml/datasets.php} for $>$ 150 data sets). In general, its objective function is fully non-separable (here we do not consider its complex dual form, which is clearly out of scope).} \cite{B_james2021introduction} for least-squares estimation. In fact, in the original Ph.D. dissertation regarding CC \cite{B_potter1997design}, all of three real-world problems considered (i.e., string cover, rule learning, and neuroevolution) are \emph{non-separable}, though CC was first benchmarked on both \emph{separable} and \emph{non-separable} artificially-constructed functions. Currently its state-of-the-art real-world applications come mainly from neuroevolution for RL (refer to Miikkulainen’s or Schmidhuber’s lab \cite{J_JMLR_gomez2008accelerated,D_EvoTorch,J_NECO_schmidhuber2007training,C_GECCO_gomez2005coevolving,C_ICML_fan2003utilizing,C_IJCAI_gomez1999solvinga,J_ML_moriarty1996efficient,C_ICML_moriarty1995efficient}). Obviously, all the loss functions used by them are \emph{non-separable,} caused by the \emph{nonlinearity} of neural network itself as well as simulation model.

Given the ubiquity of non-separable problems in practice, in this paper we focus on CC for non-separable large-scale BBO mainly from three different yet related viewpoints (i.e., problem structure, convergence analyses, and distributed acceleration).

\textbf{Problem Structure:} It is natural to deduce that not all non-separable problems can be handled efficiently by CC. For example, on many \emph{ill-conditioned} non-separable landscapes,
both CC and its gradient-based counterpart (i.e., coordinate descent, CD \cite{J_MP_wright2015coordinate}) are typically \emph{worse} than second-order-type optimizers (e.g., LM-CMA \cite{J_ECJ_loshchilov2017lmcma} and L-BFGS \cite{J_MP_liu1989limited}). On the contrary, on some other non-separable landscapes (e.g., with relatively \emph{sparse} variable interactions \cite{C_CEC_sun2019decomposition}), some modern CC could obtain very competitive (sometimes even state-of-the-art) results. Therefore, a key theoretical question arises: \emph{On what kinds of problem types from (non-separable) real-world problems CC is preferred over others?} As argued in 1989 by Conn et al. \cite{BS_conn1989introduction}, PAS functions “\emph{are clearly a very restricted case of partially separable (PS) functions}”\footnote{We notice that there are several CC papers do not distinguish PAS and PS clearly, which may cause confusions. In the mathematical optimization (MO) community, PS includes not only PAS, but also \emph{non-separable functions with a sparse linkage structure}. In effect, the focus of MO is on the latter rather than the former, as the former is seen as “\emph{a very restricted case}”, which is not emphasized in some CC papers.}. Here we exclude PAS because of its hard-to-satisfy assumption in practice.

To the best of our knowledge, there has no theoretical work to satisfactorily solve the above theoretical challenge. In this paper, we first answer a related but much simpler practical question: \emph{When may end-users prefer to use divide-and conquer methods in practice}? Some common answers are presented in the following:
\begin{itemize}
    \item There often exists a “\emph{natural}” decomposition for many complex systems: e.g., body-brain co-evolution \cite{J_RSIF_cheney2018scalable,J_Nature_lipson2000automatic}  and multi-agent learning \cite{C_AAMAS_panait2007theoretical} in the evolutionary robotics field, and expectation-maximization \cite{C_ICLR_chen2021molecule,C_NeurIPS_greff2017neural} from the AI field. Their significant feature is that all the subcomponents interact \emph{nonlinearly} but work at \emph{different time scales or computing units} (leading to different update frequencies). Note that the so-called ‘natural’ decomposition reflects the design preference or mandatory requirement (\emph{not necessarily} the ‘optimal’ decomposition solution).
    \item In parallel/distributed computing, a single computing unit (e.g., a CPU core) always has a relatively limited memory and capability with Moore Law’s ending \cite{J_Science_leiserson2020there}. For big-data driven LSO with certain structures \cite{J_SIOPT_nesterov2012efficiency,J_JMLR_richtarik2016distributed,arXiv_shi2017primer}, general decomposition strategies (e.g., CD) are a viable solution for scalability, one critical metric for any general-purpose optimizer.
    \item  There may be a relatively \emph{weak} interaction between subcomponents in some (not all) non-separable real-world optimization problems, where decomposition-based methods could converge fast (within a practically accepted accuracy). This may partly explain why there always have some researchers and practitioners using them since the establishment of the optimization area (furthermore, another advantage is their relative ease to understand and implement) \cite{J_MP_wright2015coordinate,arXiv_shi2017primer}.
\end{itemize}

\textbf{Convergence Analyses:} In order to make it mathematically tractable, we simplify CC as a continuous-game (CG) model \cite{J_TAC_ratliff2016characterization} but without losing its essential property. As compared to previous evolutionary game theory (EGT) \cite{C_AAMAS_panait2007theoretical,J_ECJ_panait2010theoretical}, our CG model can provide a \emph{much clearer} analytical perspective built on only the pure Nash equilibrium (PNE) concept, much simpler than its mixed counterparts used in EGT. We theoretically show that \emph{under what conditions} CC converges, which most convex-quadratic benchmark functions can satisfy (see Section 3). To validate its prediction ability, we further demonstrate that many \emph{overlapping} functions from the two newest test suites could \emph{essentially} satisfy these conditions, which illustrates that CC finally converges to the global optimum on them, which is proved for the first time (our simulation experiments can \emph{perfectly} match our predictions). Furthermore, using this new model, we reconfirm previously discovered pathologies of CC in a unified manner (i.e., relative generalization \cite{B_wiegand2004analysis} and loss of gradients \cite{C_PPSN_wiegand2004spatial}). Depending upon these theoretical analyses, we propose to use a \emph{hierarchical decomposition} structure to alleviate these issues and to obtain better generalizability (see Section 4.A).

\textbf{Distributed Acceleration:} Intuitively, CC appears to fit for distributed computing well \cite{C_PPSN_potter1994cooperative}. However, under nonlinear inter-dependencies between subcomponents, the parallelism of CC is still a non-trivial task, since such a nonlinearity can lead to a difficulty in credit assignment for fitness evaluations of each subpopulation. To bypass this, we use the recently proposed  \emph{multi-level learning/evolution} (MLE) framework \cite{C_PPSN_duan2022collective,J_PNAS_vanchurin2022theory}, where each subpopulation conducts (local) metric learning on only its corresponding subspace for a much lower time and space complexity, which is important to match the hierarchical memory structure of CPU. At the same time, multiple large-scale CMA-ES variants (e.g., LM-CMA) are maintained in parallel, in order to reduce the possible risk of getting trapped into a suboptimal Nash equilibrium often encountered by CC. After each relatively short learning period, all learnt information will be collected, selected, and diversified at the meta-level for next cycles. Overall, the MLE framework can combine CC's fine-tuning ability from decomposition with CMA-ES's invariance \cite{J_ASOC_hansen2011impacts} property (see Section 4).

Numerical experiments on a set of high-dimensional benchmark functions validate its search performance and scalability on an industry-level clustering computing platform with 400 cores (see Section 5).

\section{Related Works}

In this section, we first analyze the state-of-the-art CC and then discuss its real-world applications with non-separable forms and related theoretical advances. Next, we check many of \textit{partially separable} real-world problems from the mathematical optimization community, to show that nearly all of them are \textit{non-separable} according to variable interactions. Finally, following the suggestion from the latest CC survey \cite{J_TEVC_ma2019survey}, we build a connection between CC and its gradient-based counterpart (i.e., coordinate descent, CD) to help better understand the essence of decomposition-based optimizers.

\subsection{State-of-the-Art CC}

Since the publish of \cite{J_TEVC_omidvar2014cooperative}, a series of different improvements based mainly on DG (e.g., \cite{J_TEVC_vandenbergh2004cooperative, C_PPSN_chen2010largescale, J_TEVC_li2012cooperatively, J_TOMS_mei2016competitive, J_TEVC_sabar2017heterogeneous, J_TEVC_yang2017efficient, J_TEVC_omidvar2017dg2, J_TCYB_ge2017cooperative, J_TEVC_sun2018recursive, J_TCYB_ren2018boosting, J_ECJ_wang2018cooperative, J_TCYB_peng2018multimodal, J_TEVC_yang2021efficient, J_TEVC_liu2020hybrid, J_TELO_xu2021constraintobjective, J_TEVC_chen2023efficient, J_TEVC_ma2022merged, J_TEVC_wu2023cooperative, J_TEVC_kumar2022efficient, J_TEVC_xu2022difficulty}) have been proposed till now. Although these improvements often reduce the needed number of function evaluations, they might become over-skilled because they are built on only the PAS assumption\footnote{The PAS assumption may be of a \textit{theoretical} interest for some researchers.}. As stated before, \textit{what kinds of real-world applications satisfy the PAS assumption} is still an open question. Similar issue has happened in GA’s building- block hypothesis (BBH) for crossover operators (refer to \cite{J_NACO_beyer2002evolution,C_NIPS_mitchell1993when}). Like BBH, while PAS is intuitively appealing, finding functions from real-world applications to support it appeared surprisingly difficult. Although these automatic separability detection techniques sometimes could be used to analyze the variables interaction matrix as the basis of possible problem transformation for gray-box optimization \cite{J_SMO_gandomi2023variable}, such a problem transformation is unavailable in box-box scenarios. Note that the real-world problem itself in \cite{J_SMO_gandomi2023variable} is \textit{non-separable} (after transformation its new form is still \textit{non-separable}). Similarly, in both \cite{J_ECJ_wang2018cooperative} and \cite{J_TEVC_xu2022difficulty}, their tested real-world problems are also \textit{non-separable}, despite they focused on decomposition.

Till now, the decomposition of \textit{non-separable} problems is still a challenge. The above separability detection will become of less importance for non-separable functions when such \textit{a prior} knowledge (i.e., \textit{non-separability}) is relatively easy to obtain in practice (e.g., \cite{J_TEVC_chen2019cooperative,J_JMLR_gomez2008accelerated} to name but a few). Recently, Chen et al. \cite{J_TEVC_chen2022decomposition} considered \textit{non-additively partially separable} problems \cite{J_TCYB_li2023dual}. However, how to extend it for general \textit{non-separable} problems is still unclear. Komarnicki et al. \cite{J_TEVC_komarnicki2022incremental} suffered from the same issue. Zhang et al. \cite{J_TEVC_zhang2019dynamic} applied CC to \textit{non-separable} problems and obtained promising results on some benchmark functions. However, they did not analyze its convergence property. Jia et al. \cite{J_TCYB_jia2020contributionbased} extended contribution-based CC to a special type of non-separable functions (called \textit{overlapping} \cite{C_CEC_sun2019decomposition,J_TSMC_zhang2023graphbased,J_InfSci_omidvar2015designing}). Similarly, they did not analyze whether CC converges or not yet. Although recently Ge et al. \cite{J_TCYB_ge2017cooperative} presented a simple convergence analysis under the block separability assumption, it is not suitable for \textit{non-separability}.

\subsection{Theoretical Advances of CC}

There have been a series of theoretical works based on EGT to analyze complex convergence behaviors of CC on non-separable functions. Wiegand \cite{B_wiegand2004analysis} for the first time proved that its replicator equations converge to PNE and showed that CC sometimes suffers from the \textit{relative generalization} issue. Although the following-up works (e.g., \cite{J_ECJ_panait2010theoretical,C_FOGA_wiegand2002modeling,J_TEVC_ficici2005gametheoretic}) extended Wiegand’s work somewhat, they are not extended to current large-scale CC versions due to the following factors: 1) It considers only one very simple decomposition case on a (discretized) low-dimensional search space; 2) it depends on a highly simplified payoff matrix for computing mixed Nash equilibria and therefore it cannot consider complex fitness landscape explicitly; 3) its derived lenient learning is still rarely used for LSO \cite{C_AAMAS_panait2007theoretical}. Overall, although they could provide a valuable analytical tool for CC, new theoretical advances are still expected for LSO-focused CC.

In \cite{C_GECCO_jansen2003exploring} and \cite{J_ECJ_jansen2004cooperative}, Jansen and Wiegand provided the first expected runtime analysis for CC. Surprisingly, they found that “\textit{the property of separability is neither a sufficient one to imply an advantage of CC, nor is inseparability a sufficient enough property to imply a disadvantage}”. Similarly, Gomez et al. \cite{J_JMLR_gomez2008accelerated} argued that “\textit{much of the motivation for using the CC approach is based on the intuition that many problems may be decomposable into weakly coupled low-dimensional subspaces that can be searched semi-independently by separate species. Our experience shows that there may be another, complementary, explanation as to why cooperative coevolution in many cases outperforms single-population algorithms}”. Overall, a deeper convergence analysis is highly desirable for recent large-scale CC versions.

\subsection{Partially Separability and Coordinate Descent}

Originally, the concept of partially separability (PS) \cite{R_toint1984test} was defined in 1981 from the mathematical optimization field and is still studied by its one original author Toint now \cite{J_TOMS_porcelli2022exploiting}. Until 2010, its very special form (i.e., PAS) was popularized for CC in the large-scale BBO field via a series of IEEE-LSGO competitions, despite there was one earlier (1999) EA paper also involving it \cite{J_ECJ_muhlenbein1999fdaa}. Here, we re-check PS based on the original paper\footnote{The first (1981) paper of partially separability is not accessible online. Unfortunately, the original author (Toint) cannot provide its pdf version (via email communication). Instead, we use his 1983 version, which provides a set of 50 functions. Again, this 1983 paper also becomes inaccessible online now (to our knowledge). Luckily, we have saved it locally before (if you want to read it, please contact Toint or us only for academic purpose).}. Surprisingly, there is \textbf{only one} function in \cite{R_toint1984test} that meets the PAS assumption in all 43 functions (here we have excluded 7 extra low-dimensional functions, because they are only used to test programming correctness). More et al. \cite{R_averick1992minpack2} collected a total of 24 optimization problems from many real-world applications. \textbf{All\footnote{We have excluded 4 problems since their objective function formula are not given \textit{explicitly}, though they seems to be also non-separable.} of} them have a \textit{non-separable} form \cite{J_COA_bouaricha1997impact}. Arguably, most PS instances from the mathematical optimization field are \textit{non-separable}, obviously different from previous IEEE-LSGO competitions. We notice that some of recent CC papers start to extend previous IEEE-LSGO competitions to non-separable cases via e.g., overlapping functions. Similarly, coordinate descent (CD) is commonly used to optimize non-separable problems (often with friendly structures), since its first application in 1954 \cite{J_JASA_hildreth1955corrigenda}. Note that for CD separability is limited to only the regularization term (in other words, the whole function form is generally inseparable).

\section{Convergence Analyses via a Continuous-Game Model}
\label{sec:continuous_games}

In this section, we propose a continuous-game model as a basis of convergence analyses of CC. To validate its prediction ability, the convergence behaviors of CC are analyzed on common fitness models and the latest \textit{overlapping} functions.

\subsection{Continuous-Game (CG) Model of CC}

For any objective function\footnote{Only the \textit{minimization} of objective (fitness) functions is considered, since maximization can be transformed into minimization simply by negating it.} $f(\mathbf{x}): \mathbb{R}^n \rightarrow \mathbb{R}$, CC divides all its coordinate indexes $\{1, \ldots, n\}$ into a set of mutually exclusive partitioning groups $p=\left\{g_1, \ldots, g_m\right\}$ where $1<m \leq n$ and $g_i \neq \emptyset, g_i \cap g_{j \neq i}=\emptyset, \cup_i g_i=\{1, \ldots, n\}, \forall i, j \in\{1, \ldots, m\}$. Note that $1<m$ excludes no decomposition case. An interesting theoretical question is \textit{how many} possible partitions exist for decomposition-based methods. The Bell number, a computing concept from \textbf{combinatorial mathematics}, can help answer this question. For example, even for a 25-d function, there are totally 4,638,590,332,229,999,352 partitioning solutions (even when the partitioning order is not considered here) \cite{D_A000110}.

\subsubsection{Ubiquity of Pure Nash Equilibrium (PNE)}

In game theory, it is well-known that there \textit{always} exists at least one \textit{mixed} (not necessarily \textit{pure}) Nash equilibrium for any noncooperative game \cite{J_PNAS_nash1950equilibrium, J_AnnMath_nash1951noncooperative}. However, our previous conference paper has mathematically shown that CG’s special payoff structure guarantees the ubiquity of the pure Nash equilibrium (PNE) for \textit{any} one partitioning (refer to \cite{C_CEC_duan2019when} for details). Hereinafter, in order to avoid ambiguity, we will focus on only the PNE as defined below.

\begin{figure}[htbp]
\centering
\begin{minipage}{0.49\linewidth}
    \centering
    \includegraphics[width=1\linewidth]{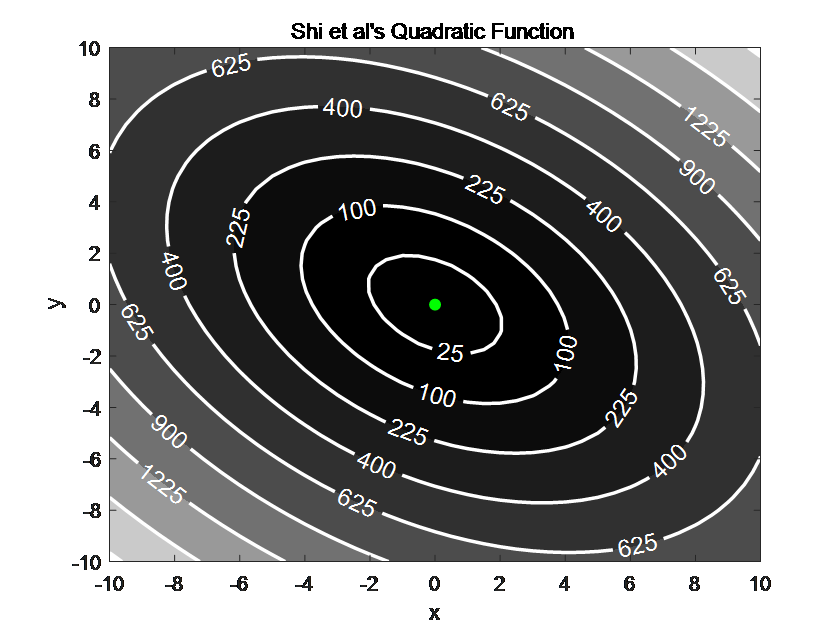}
    \subcaption{$f_1(x,y)=7x^2+6xy+8y^2$}
    \label{fig:visualization_nash_1}
\end{minipage}
\begin{minipage}{0.49\linewidth}
    \centering
    \includegraphics[width=1\linewidth]{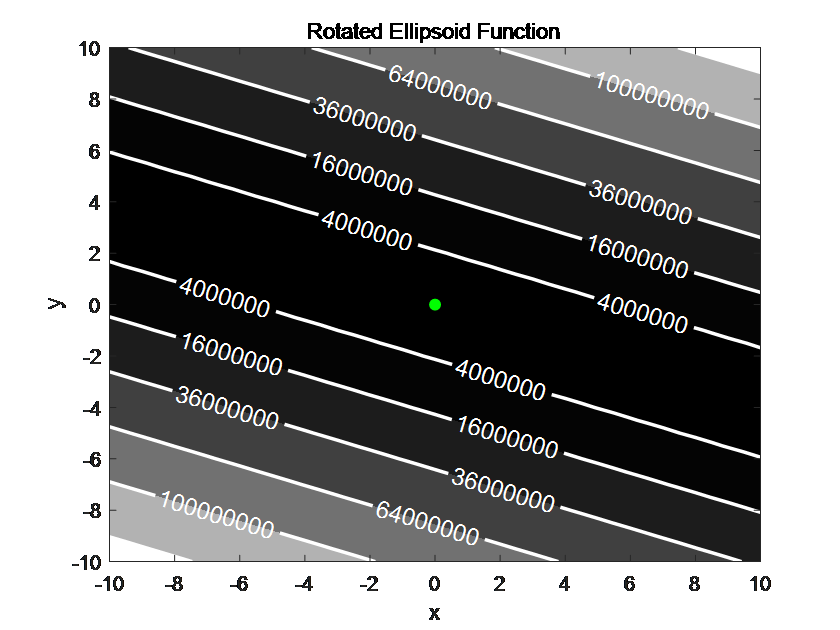}
    \subcaption{$f_2(x,y)=x^2+10y^2$ (\textbf{rotated})}
    \label{fig:visualization_nash_2}
\end{minipage}
\begin{minipage}{0.49\linewidth}
    \centering
    \includegraphics[width=1\linewidth]{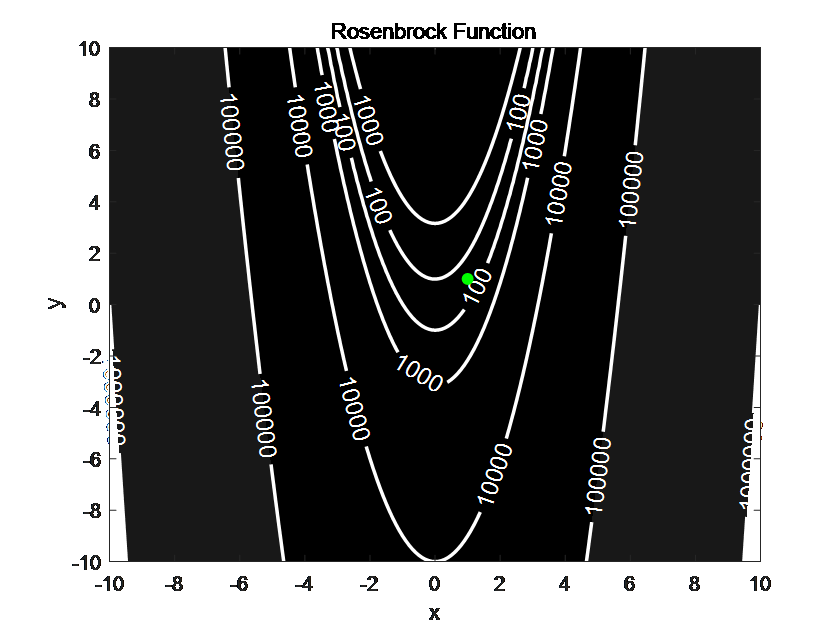}
    \subcaption{$f_3(x,y)=100(x^2-y)+(x-1)^2$}
    \label{fig:visualization_nash_3}
\end{minipage}
\begin{minipage}{0.49\linewidth}
    \centering
    \includegraphics[width=1\linewidth]{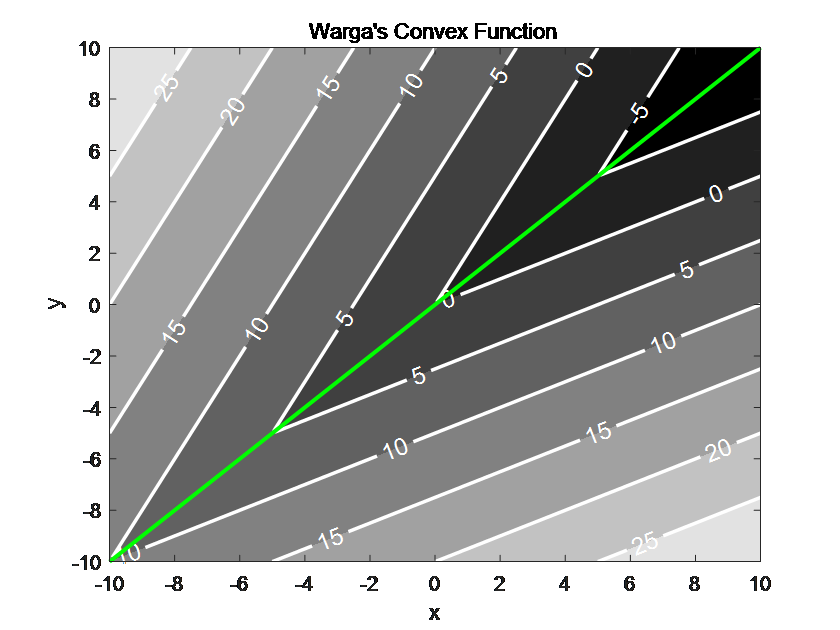}
    \subcaption{$f_4(x,y)=|x-y|-\min(x,y)$}
    \label{fig:visualization_nash_4}
\end{minipage}
\qquad
\caption{Visualization of all pure Nash equilibria (PNE) on four 2-d non-separable
functions. The white line represents the contour levels of fitness landscape
while the green point/line (aka best-response curve) denotes the location(s) of PNE.}
\label{fig:visualization_nash}
\end{figure}

\textit{Definition} \textbf{(Pure Nash Equilibrium, PNE)}. Given \textit{any} partition $p=\left\{g_1, \ldots, g_m\right\}$ of the fitness function $f(\mathbf{x}): \mathbb{R}^n \rightarrow \mathbb{R}$, we say that its decision vector $\mathbf{x} \in \mathbb{R}^n$ is one \textit{pure Nash equilibrium} (w.r.t. $p$ ) iif the decision subvector $\mathbf{x}_{g_i} \in \mathbb{R}^{\left|g_i\right|}$ for each $g_i \in$ $p, \forall i \in\{1, \ldots, m\}$, satisfies
\begin{equation}
f\left(\mathbf{x}_{g_i}, \mathbf{x}_{\neq g_i}\right) \leq f\left(\mathbf{x}_{g_i}^{\sim}, \mathbf{x}_{\neq g_i}\right), \forall \mathbf{x}_{g_i}^{\sim} \in \mathbb{R}^{\left|g_i\right|} \backslash\left\{\mathbf{x}_{g_i}\right\},\label{eq:pure_nash}
\end{equation}
where $\mathbf{x}_{\neq g_i}$ denotes all the remaining decision subvectors excluding $\mathbf{x}_{g_i}$ and $\mathbf{x}_{g_i}^{\sim}$ denotes any other possible values (i.e., $\left.\mathbb{R}^{\left|g_i\right|} \backslash\left\{\mathbf{x}_{g_i}\right\}\right)$. If (\ref{eq:pure_nash}) is strict, we say that $\mathbf{x}$ is a \textit{strict} PNE (w.r.t. $p$ ). Since this concept relies heavily on the pre-partitioning, we add a suffix “(w.r.t. $p$)” to emphasize and clarify it, if necessary. To help understand this solution concept for decomposition-based optimization, we visualize its distributions on four common \textit{non-separable} test functions in Fig. \ref{fig:visualization_nash}.

As is seen in Fig. \ref{fig:visualization_nash}, the top-left is a \textit{strictly-convex} function, designed by Shi et al. \cite{arXiv_shi2017primer} from the latest CD survey. There is a unique PNE, which is the global optimum (CC could converge to it rapidly owing to weak dependency). The top-right is an \textit{ill-conditioned} (rotated) Ellipsoid function commonly used to benchmark the local search ability of EAs. Similarly, there is only one PNE (CC still could converge to it but its convergence rate is very slow owing to strong dependency). The bottom-left is a very famous \textit{non-convex} function \cite{J_ECJ_shang2006note}, where there is also a unique PNE (CC still could converge to it but the convergence rate is slow when approaching the parabolic curve). The bottom-right is a \textit{convex but non-differentiable} function, originally proposed by Warga \cite{J_JSIAM_warga1963minimizing} from the CD community. For it, there is a \textit{continuum} of PNE (CC does not necessarily converge to the global optimum but to other suboptimal PNE, depending on the starting point). For their convergence demonstrations, please refer to \ref{appendix:convergence_four}.

\textit{Lemma 1}: Given \textit{any} partitioning solution $p=\left\{g_1, \ldots, g_m\right\}$ of $f(\mathbf{x})$, its global optimum $\mathbf{x}^*$ is a pure Nash equilibrium w.r.t. $p$. However, the inverse is \textit{not necessarily} true.

The ubiquity of PNE under \textit{any} partition can be guaranteed by the above lemma, only if the objective function has (at least) one global optimum (which is a default assumption in most theoretical analyses).

\textit{Lemma 2}: Given \textit{any} partition $p=\left\{g_1, \ldots, g_m\right\}$ of $f(\mathbf{x})$ : $\mathbb{R}^{n>2} \rightarrow \mathbb{R}$ where $\left|g_i\right|>1, \exists i \in\{1, \ldots, m\}$, we can divide $p$ along $g_i$ and get a new partition $p^{\prime}$ satisfying $\left|p^{\prime}\right|>|p|$. If $x$ is a PNE w.r.t. $p$, then $\mathbf{x}$ is also a PNE w.r.t. $p^{\prime}$. However, the inverse does not necessarily hold. (We term this interesting property as the \textit{\textbf{downward rather upward propagation}} of $P N E$.)

\begin{figure}[htbp]
    \centering
    \includegraphics[width=0.8\linewidth]{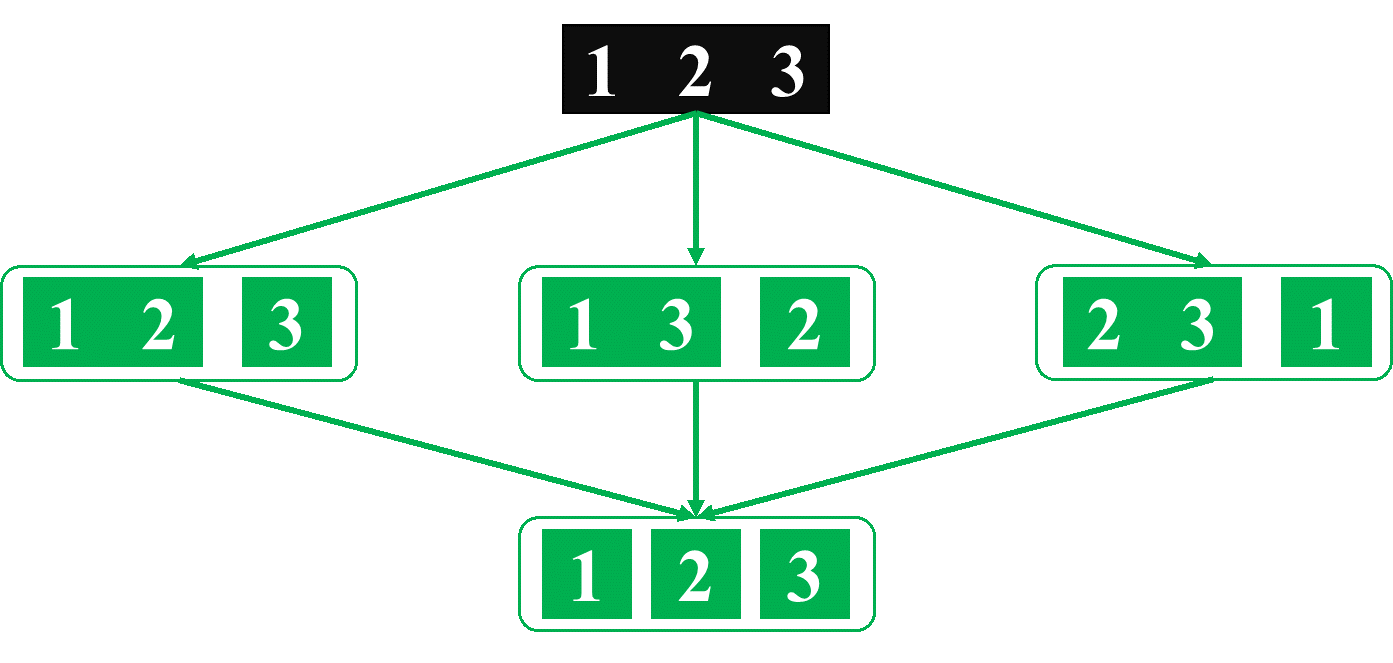}
    \caption{A simple example to illustrate the \textit{downward rather upward
propagation} of PNE along hierarchical partitions.}
    \label{fig:downward_propagation}
\end{figure}

Since \textit{lemma 2} plays a fundamental role in the following convergence analyses, we design a simple example to illustrate it. Consider an artificially designed function case $f(x, y, z)=$ $f(x, y)+f(y, z)$, where $f(x, y)=(x+y)^2+(y-1)^2$ and $f(y, z)=(y+1)^2+(y+z)^2$. For it, there are 4 partitions: $\{\{1,2\},\{3\}\},\{\{1,3\},\{2\}\},\{\{2,3\},\{1\}\}$, and $\{\{1\},\{2\},\{3\}\}$, as shown in Fig. \ref{fig:downward_propagation} (for simplicity, the order of partitions is not considered here). Interestingly, all partitions can be well reorganized in a \textit{hierarchical} (\textit{recursive}) manner. Each green arrow represents one further partition operation (i.e., from $p$ to $p^{\prime}$ ). The global optima, a special type of PNE, can propagate both \textit{downwardly} (from $p$ to $p^{\prime}$ ) and \textit{upwardly} (from $p^{\prime}$ to $p$ ). One theoretical significance of \textit{lemma 2} is that it can greatly simplify the difficult seeking process of PNE, as proved in the next subsection.

In a nutshell, all global optima are PNE; however, the inverse does not necessarily hold (as in Fig. \ref{fig:visualization_nash}, we give a counterexample to prove it). Unfortunately, there is no certain mapping relationship between PNE and local optima: Some local optima may be PNE but some may be not; similarity some PNE may be local optima while some may be not (please refer to Fig. 1 for a counterexample). Here we choose to omit their rigorous proofs owing to their mathematical ease to design counterexamples. Note that since all local optima are global optima for convex functions, this greatly simplifies the global convergence analyses of CC on this specific yet important function form (see the Main Theorem for proof details).

\subsubsection{Convergence to a PNE in CG}

Although it is well-known in the scientific community that “\textit{essentially all models are wrong}” \cite{J_ECJ_panait2010theoretical}, a good model for CC should meet (at least) the following two criteria while omitting the peripheral details: 1) It should still capture the essence of the algorithm even after simplifications, which can be used to predict the \textit{limit} convergence behavior; 2) It should exhibit somewhat \textit{prediction and even extrapolation} ability, despite it is nearly impossible to exclude the failure risk on some unseen practical scenarios.

In previous EGT models \cite{J_ECJ_panait2010theoretical,B_wiegand2004analysis}, CC was modeled using two-player replicator equations. Alternatively, CC can also be simply modeled as a continuous game (CG) on the original continuous search space, which is comprised of multiple dynamically-coupled subpopulations. Continuous games have been investigated in many research areas (economics \cite{J_ECMA_rosen1965existence}, nonlinear automatic control \cite{J_TAC_tatarenko2018learning,J_TAC_ratliff2016characterization}, multi-agent learning \cite{C_NeurIPS_lanctot2017unified}, engineering design \cite{B_hespanha2017noncooperative}, and AI \cite{C_NIPS_goodfellow2014generative,J_NN_schmidhuber2020generative}, etc.).

In principle, our CG model can be seen as a \textit{particular} form of CG proposed by Ratliff et al. \cite{J_TAC_ratliff2016characterization} from the automatic control community. For the former, all the players have the same objective function form but with a disjoint subset of decision vectors. However, for the latter, each player has its own objective function, which generally has a different form with each other. We borrow the classical model \cite{J_NRL_hildreth1957quadratic,J_NRL_desopo1959convex} from CD to formalize each cycle of CC, mathematically represented as one \textit{\textbf{transformation}} $T: \Omega \subset \mathbb{R}^n \rightarrow \Omega \subset \mathbb{R}^n$ of the search space $\Omega$ into $\Omega$ itself, satisfying the two following conditions:
\begin{enumerate}
    \item $f(T(\mathbf{x})) \leq f(\mathbf{x})$, where $\mathbf{x}$ is the best-so-far solution;
    \item $f(T(\mathbf{x}))=f(\mathbf{x}) \Leftrightarrow T(\mathbf{x})=\mathbf{x}$. (a fixed point)
\end{enumerate}

Condition 2 excludes one very special case (that is, when $f(T(x))=f(\mathbf{x}), \mathbf{T}(\mathbf{x}) \neq \mathbf{x})$, which otherwise results in the \textit{cyclical} behavior of the best-so-far solution and prevents the global convergence (which was found first by Powell in 1973 \cite{J_MP_powell1973search}).

For successive transformation processes, the key point is to extract a \textit{strictly decreasing} fitness subsequence from the nonincreasing fitness sequence of the best-so-far solution $\mathbf{x}$ generated by $\mathrm{CC}$. This could be guaranteed by a strong assumption that each suboptimizer has \textit{sufficient} search ability for each subproblem (that is, \textit{it can find the global minimizer for each subproblem given a finite number of iterations}). It is worthwhile noting that even such a strong assumption cannot ensure the convergence to the global optimum on the original problem (see Fig. \ref{fig:visualization_nash_4} for an example). As a result, finally $\lim _{t \rightarrow+\infty} T_t(\mathbf{x})$ will move towards \textbf{a fixed point} (that is, ${T}(\mathbf{x})=\mathbf{x})$, which is also a PNE\footnote{Its mathematical proof is simple via proof by contradiction: In fact, the Nobel-prize winner Nash was the first to connect the fixed point with the equilibrium point (now named after him) in his seminar \textit{PNAS} and \textit{Annals of Mathematics} papers, laying the theoretical foundation of modern game theory.}. Please refer to \ref{appedix:convergence_pne} for a detailed mathematical demonstration. In practice, the global minimizer of each subproblem in inner cycle is not needed to find exactly \cite{C_CEC_sun2019decomposition} in outer cycle.

We acknowledge that the above CG model is very simple from a practical view of point. Based on it, however, we will derive some interesting theoretical results, as presented below.

\subsection{Convergence Analyses on Convex Functions}

Based on the above CG model, the convergence problem of CC can be well converted to the \textit{seeking} and \textit{classification} problem of PNE. Here we provide a main theorem on convex functions, as presented below:

{\textbf{Main Theorem}\label{main_theorem}}: Consider any partitioning solution $p=\left\{g_1, \ldots, g_m\right\}$ of a continuous-differentiable convex function $f(\mathbf{x}) \in {C}^1(\Omega, \mathbb{R})$, where $\Omega$ is an open convex set\footnote{Here it is implicitly assumed that there is (at least) one global optimum in this open convex set.} in $\mathbb{R}^n$. For $f(\mathbf{x})$, all PNE (w.r.t. $p$) are equivalent to the global optima, and vice versa.

Proof: First we demonstrate one special partition case $p^n=$ $\{\{1\}, \ldots,\{n\}\}$ (i.e., $m=n$ ). Assume that $\widetilde{\mathbf{x}}$ is a PNE (w.r.t. $p^n$). By the definition of PNE, we can simply infer that $\widetilde{\mathbf{x}}_{\{i\}} \in$ $\underset{x_{\{i\}}}{{argmin}} f\left(\mathbf{x}_{\{i\}}, \mathbf{x}_{\neq\{i\}}\right) \subseteq \Omega, \forall i \in\{1, \ldots, n\}$.

Owing to the continuous differentiability of $f(x)$, each partial derivative $\nabla_{\{i\}} f(\widetilde{x})=0, \forall i \in\{1, \ldots, n\}$. So, $\tilde{x}$ is also a stationary point. Note that for $\underset{x_{(i)}}{{argmin}} f\left(\mathbf{x}_{\{i\}}, \mathbf{x}_{\neq\{i\}}\right)$, the solution is not necessarily unique. According to the convexity of $f(\mathbf{x})$, it is well-known that all stationary points are the global optima. With \textit{lemma 1} (that is, all global optima are PNE), we conclude that for this special case $p^n=\{\{1\}, \ldots,\{n\}\}$, all PNE (w.r.t. $p^n$) are the global optima, and vice versa.

Then, we show the general case for any other partition $p=$ $\left\{g_1, \ldots, g_m\right\}$ when $m \neq n$. Based on \textit{downward propagation} of PNE (\textit{lemma 2}), all PNE w.r.t. $p$ belongs to a set of PNE w.r.t. $p^n$, which are also global optima. With \textit{lemma 1} , we finish the proof perfectly.

Clearly the above main theorem is valuable to understand why CC could converge to the global optima on many (though not all) non-separable functions. In effect, many of common test functions fall into this interesting class \cite{J_AIJ_whitley1996evaluating}. Take e.g., \textit{Cigar}, \textit{Discus}, \textit{CigarDiscus}, \textit{Ellipsoid}, and \textit{Schwefel} as examples for convex-quadratic functions \cite{J_OMS_hansen2021coco}. Note that their \textit{non-separable} rotated-and-shifted versions still fall into this class. Surprisingly, all the overlapping functions from the newest LSO test suite for CC \cite{C_CEC_sun2019decomposition} essentially still meet these above assumptions (after suitable simplifications). The above main theorem can in part explain why one state-of-the-art CC version could obtain very competitive performance on a particular class of overlapping functions, despite all of them are non-separable (refer to the following subsection for more details). In most of previous CC papers for LSO, both the theoretical and practical significance of PNE are overlooked or blurred. However, in the non-separable cases, there are more complex relationships between PNE and global optima, which need to be clarified theoretically \cite{J_TEVC_wolpert1997no}.

\subsection{Global Convergence Guarantee on Overlapping Functions}

In order to validate the \textit{extrapolation} ability of our proposed CG model, here we consider a class of \textit{overlapping} (a special type of non-separability) functions from the latest LSO test suite \cite{C_CEC_sun2019decomposition}. The original experiments reported in \cite{C_CEC_sun2019decomposition} showed that on each function the best-so-far solution is far from the global optimum, since a \textit{relatively small} maximum of function evaluations (i.e., 3e6) was used. Based upon \textit{corollary 1} shown below, however, we predict that CC could converge to the global optimum on them finally, since they \textit{essentially} meet the assumptions of the main theorem, when two benchmarking operations (i.e., non-smoothing and asymmetry) are removed to make the involved mathematics tractable. To validate our prediction, we increase the maximum of function evaluations from 3e6 to 3e7 significantly. New experiment results are in accordance with our theoretical prediction\footnote{The source code is freely available at \url{https://bitbucket.org/yuans/rdg3.}}(even when two benchmarking operations are added).

\textit{Corollary 1}: Consider a non-separable overlapping function $f(\mathbf{x})=$ $\sum_{i=1}^m f_i\left(\mathbf{x}_i\right)$ with $\mathbf{x} \in \Omega \subseteq \mathbb{R}^n, \mathbf{x}_i \cap \mathbf{x}_j \neq \emptyset, \exists i \neq j \in$ $\{1, \ldots, m\}$ and $U_i \mathbf{x}_i=\mathbf{x}$. All subfunctions $f_i$ have the exactly same\footnote{In effect, such a condition can be further weakened, only if it is convex and continuous-differentiable. For simplicity, however, we exclude it here.} base form $f_b(\mathbf{x})=\sum_{i=1}^n\left(\sum_{j=1}^i \mathbf{x}_i\right)^2$ (i.e., \textit{Schwefel's Problem 1.2} in \cite{C_CEC_sun2019decomposition}). Its rotated-and-shifted version $f_s^r(\mathbf{x})=$ $f_b(\mathbf{R}(\mathbf{x}-\mathbf{s}))$ is used by each $f_i\left(\mathbf{x}_{\mathbf{i}}\right)$, where $\mathbf{R}$ is an orthogonal matrix and $s$ is a shift vector. For this class of overlapping functions, all PNE (w.r.t. \textit{any partition}) are global optima, and vice versa.

Proof: First we show the each continuous-differentiable subfunction $f_i, \forall i \in\{1, \ldots, m\}$, is convex. Since all $f_i$ share the same form $f_b(\mathbf{x})=\sum_{i=1}^n\left(\sum_{j=1}^i \mathbf{x}_j\right)^2$, we only need to prove the base form case. Its Hessian $\mathbf{H}$ is
\[
2 *\left[\begin{array}{cccccc}
n & n-1 & n-2 & & 2 & 1 \\
n-1 & n-1 & n-2 & \cdots & 2 & 1 \\
n-2 & n-2 & n-2 & & 2 & 1 \\
& \vdots & & \ddots & \vdots \\
2 & 2 & 2 & \cdots & 2 & 1 \\
1 & 1 & 1 & & 1 & 1
\end{array}\right]
\]

Successively adding a multiple of row $i-1$ to another row $i$ by $-\frac{n-(i-1)}{n-(i-2)}, i=n, n-1, n-2, \ldots, 3,2$, we get a new matrix:
\[
\mathbf{A}=2 *\left[\begin{array}{cccccc}
n & n-1 & n-2 & & 2 & 1 \\
0 & a_{22} & a_{23} & \cdots & a_{2(n-1)} & a_{2 n} \\
0 & 0 & a_{33} & & a_{3(n-1)} & a_{3 n} \\
& \vdots & & \ddots & \vdots & \\
0 & 0 & 0 & \ldots & a_{(n-1)(n-1)} & a_{(n-1) n} \\
0 & 0 & 0 & & 0 & 1 / 2
\end{array}\right],
\]
where $a_{i i}=(n-(i-1))-\frac{n-(i-1)}{n-(i-2)} *(n-(i-1))>0, i=$ $n, n-1, n-2, \ldots, 3,2$. By the theorems from linear algebra (pp. 60-61 in \cite{B_leon2010linear}), $\mathbf{H}$ is row equivalent to $\mathbf{A}$. With the theorem from linear algebra (pp. 370 in \cite{B_leon2010linear}), we conclude that $\mathbf{H}$ is a \textit{positive definite} matrix, which means that $f_b(\mathbf{x})$ is a (strictly) convex function.

Second, we consider its rotated-and-shifted variant $f_s^r(\mathbf{x})=$ $f_b(\mathbf{R}(\mathbf{x}-\mathbf{s}))$, where $\mathbf{R}(\mathbf{x}-\mathbf{s})$ is still a convex set. For $\forall \mathbf{x} \neq$ $\mathbf{y} \in \Omega$ and $\theta \in[0,1], f_s^r(\theta \mathbf{x}+(1-\theta) \mathbf{y})=f_b(\mathbf{R}(\theta \mathbf{x}+$ $(1-\theta) \mathbf{y})-\mathbf{s}))=f_b(\theta \mathbf{R}(\mathbf{x}-\mathbf{s})+(1-\theta) \mathbf{R}(\mathbf{y}-\mathbf{s}))$, by the convexity of $f_b, \leq \theta f_b(\mathbf{R}(\mathbf{x}-\mathbf{s}))+(1-\theta) f_b(\mathbf{R}(\mathbf{y}-\mathbf{s}))=\theta f_s^r(\mathbf{x})+(1-\theta) f_s^r(\mathbf{y})$. Therefore, its rotated-and-shifted variant is still a convex function.

Then, if $f_i(\mathbf{x}), \forall i=1 \ldots, m$, is a convex function, their nonnegative weighted sum is still a convex function (pp. 79 in \cite{B_boyd2004convex}). A total of 20 overlapping functions in \cite{C_CEC_sun2019decomposition}, no matter with \textit{conforming} or \textit{conflicting} components, still meet it. With the main theorem, we can conclude the proof.

Considering that recent $\mathrm{CC}$ works focus on the non-separable overlapping cases, the above corollary can provide a \textit{relatively deep} theoretical understanding to CC's convergence properties on some overlapping functions with certain structures. Interestingly, the above corollary can also be extended to some overlapping functions designed in \cite{J_TCYB_jia2020contributionbased}, since their base form is also convex.

\section{Distributed Multi-level Learning}

In this section, a distributed cooperative coevolution (DCC) framework is proposed where the recent multilevel learning model for biological evolution 
 \cite{J_PNAS_vanchurin2022theory,C_PPSN_duan2022collective} is employed, in order to improve the efficiency and effectiveness of CC on modern clustering computing platforms. Given the high complexity of distributed algorithms, we provide its source code available at GitHub\footnote{\url{https://github.com/Evolutionary-Intelligence/DCC}} for our proposed framework to ensure repeatability. For distributed algorithms, multiple performance trade-offs must be considered: such as, distributed scheduling, distributed (shared) memory management, network communications, fault tolerant, system control, and so on. In this paper, we focus on the \textit{application-level} parallelism and leave other tedious tasks to the underlying distributed computing engine.

\subsection{Hierarchical Structure of DCC}

\begin{algorithm}[!t]
\caption{Distributed Cooperative Co-evolution (DCC) with CMA-based Multilevel Learning. (Refer to its open-source Python code for repeatability: \url{https://github.com/Evolutionary-Intelligence/DCC}.)}\label{alg:alg1}
\begin{algorithmic}[1]
\Require{\(p\): total number of available slave nodes in computing platform}
\Statex{\(p\_es\): number of slave nodes to run LM-CMA}
\Statex{\(p\_cc\): number of slave nodes to run CC}

\Ensure{\(b\_x\): the best-so-far solution}
\Statex{\(b\_y\): the best-so-far fitness}

\State{\textbf{Initialize}: \(b\_x \gets \infty, b\_y \gets \infty\)}
\Statex{~~~~~~~~~~~~~~~randomly initialize LM-CMA and CMA-ES in CC} \Comment{\textcolor{red}{parallel}}

\While{all termination conditions are not satisfied}

\For{\(i=1\) to \(p\)} \Comment{in parallel}
    \If{\(i \leq p\_es\)} \Comment{for LM-CMA}
        \State{use Meta-ES to set global step-size}
        \State{use elitist or weighted averaging to set distribution mean}
        \State{run serial LM-CMA in original search space}
    \Else \Comment{for CMA-ES in \(p\_cc\) CC}
        \State{decompose original space into \(k\) subspaces}
        \For{\(d\) from 1 to \(k\)} \Comment{in serial for each CC}
            \State{run serial CMA-ES in \(d\)-th subspace } \Comment{in low dimensions}
        \EndFor
    \EndIf
\EndFor

\State{synchronize results from all optimizers} \Comment{expensive operation}
\LComment{the following three parts are executed serially}
\State{update distribution mean via weighted averaging at meta-level}
\State{conduct collective learning of covariance matrices} \Comment{meta-diversity}
\State{update \(b\_x, b\_y\) if a better solution is found}

\EndWhile
\end{algorithmic}
\end{algorithm}

\begin{figure}[htbp]
    \centering
    \includegraphics[width=0.9\linewidth]{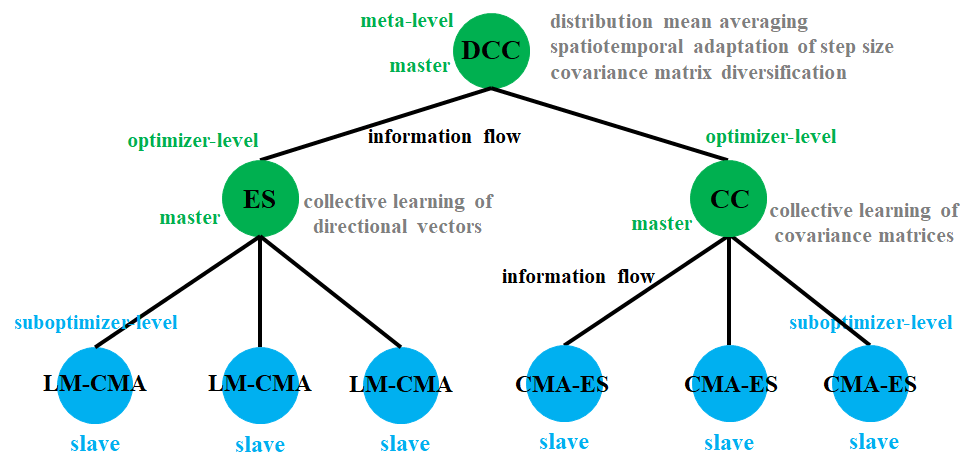}
    \caption{A hierarchical structure of our distributed cooperative coevolution framework (DCC) for large-scale BBO.}
    \label{fig:hierarchical_structure}
\end{figure}

As previously shown, CC tends to convergence to the PNE. If the PNE is not the global optimum (refer to Fig. \ref{fig:visualization_nash} as an example), CC easily gets trapped into this suboptimal solution. This issue (known as \textit{relative generalization}) is attributed to decomposition, which cannot be avoided in essence but may be alleviated somewhat. On a class of \textit{ill-conditioned} non-separable functions \cite{J_TEVC_vandenbergh2004cooperative}, like its gradient-based counterpart (CD), CC often shows \textit{much slower} convergence rates than second-order-type optimizers (e.g., LM-CMA and L-BFGS). However, when as a \textit{general-purpose} BBO framework for LSO, it is highly desirable that CC could also handle these challenging issues. To achieve this goal, we use the recently proposed multilevel learning framework \cite{J_PNAS_vanchurin2022theory,C_PPSN_duan2022collective} to combine the best of both worlds (i.e., the powerful invariance property of LM-CMA and the fine-tuning ability of CC via CMA-ES on multiple low-dimensional subspace). In this paper, we only consider using the hierarchical structure to implement it for simplicity, as seen in Fig. \ref{fig:hierarchical_structure}. In principle, a more general \textit{recursive} structure can be also applied here, which we leave for a future work.

As shown in Fig. \ref{fig:hierarchical_structure}, there are multiple slave nodes as well as one master node for distributed computing (e.g., on Spark \cite{J_CACM_zaharia2016apache} and Ray \cite{C_OSDI_moritz2018ray}). For DCC, at the meta-level the master node mainly coordinates different optimizers, each of which is run in one separate computing unit (i.e., a single CPU core). Here we choose two different ES versions (CMA-ES \cite{J_JMLR_ollivier2017informationgeometric} and LM-CMA \cite{J_ECJ_loshchilov2017lmcma}) as two search engines for the low-dimensional (often $<50$) subspace (after decomposition) and the original large-scale ($\gg 100$) space, respectively. In principle, any other EAs (e.g., GA \cite{J_JACM_holland1962outline}, EP \cite{BS_fogel1999overview}, PSO \cite{C_ICNN_kennedy1995particle}, DE \cite{J_JGO_storn1997differential}, and EDA \cite{B_larranaga2002estimation}) can be also selected as the suboptimizer. The rationale behind using two search engines lies that in the \textit{original} space LM-CMA has a high possibility to escape from the suboptimal PNE via learning of promising evolution paths while maintaining a low computationally complexity (see \cite{J_ECJ_loshchilov2017lmcma} for analysis) and in the \textit{decomposed} subspace the full-fledged CMA-ES \cite{J_JMLR_ollivier2017informationgeometric} can well approximate the inverse of its Hessian matrix for fine-tuning with a reasonable memory consumption on each computing unit (which is vital for scalability of distributed algorithms for LSO). Refer to the following three subsections as well as Algorithm \ref{alg:alg1} for more details.

\subsection{Collective Learning of Covariance Matrices}

For both the LM-CMA and the CMA-ES, covariance matrix adaptation (CMA) plays a central role in their invariance property. Owing to its cubic (or quadratic after modifications) time complexity, CMA is difficult to directly apply to LSO. Instead, approximating CMA with low-memory or low-rank techniques is more acceptable for LSO, resulting in lower time and space complexities (e.g., $O(n*\log(n))$ in \cite{J_ECJ_loshchilov2017lmcma}). For instance, CMA’s \textit{rank-one} update (without the \textit{rank}-$\mu$ update) can be finally transformed to the following nonrecursive form for offspring sampling (refer to \cite{J_TEVC_loshchilov2019large} for a detailed mathematical deduction):
$$
\begin{aligned}
\mathbf{d}^t= & \left(\left(1-\frac{c_1}{2}\right) \mathbf{I}+\frac{c_1}{2} \mathbf{p}^1\left(\mathbf{p}^1\right)^T\right) * \ldots \\
& *\left(\left(1-\frac{c_1}{2}\right) \mathbf{I}+\frac{c_1}{2} \mathbf{p}^{t-1}\left(\mathbf{p}^{t-1}\right)^T\right) \\
& *\left(\left(1-\frac{c_1}{2}\right) \mathbf{I}+\frac{c_1}{2} \mathbf{p}^t\left(\mathbf{p}^t\right)^T\right) \mathbf{z}^t,
\end{aligned}
$$
where $\mathbf{d}^t \in \mathbb{R}^n, \mathbf{I} \in \mathbb{R}^{n * n}, \mathbf{p}^t \in \mathbb{R}^n, \mathbf{z}^t \in \mathbb{R}^n \sim N(\mathbf{0}, \mathbf{I}), 0<c_1$ $<1$ are the directional vector, identity matrix, evolution path, standard Gaussian permutation, and learning rate of the rank-one update for iteration $t$, respectively. In the actual code implementation, $\mathbf{I}$ is omitted since $\mathbf{I}\mathbf{z}^t = \mathbf{z}^t$ . The above form should be implemented from right to left, in order to avoid the computationally expensive matrix operation. Only $m = O(\log(n)) \ll n$ dot products (with $O(n*\log(n))$ computational complexity) are involved when sampling each offspring.

Although they fit for distributed computing owing to their low memory requirements, these approximations may lose sufficient diversity on the covariance matrix: 1) the \textit{rank}-$\mu$ update is not used, which is beneficial for rugged fitness landscapes; 2) the \textit{rank-one} update is mainly beneficial for a predominated search direction (e.g., \textit{Cigar} and \textit{Rosenbrock} \cite{J_OMS_hansen2021coco}) rather than multiple promising search directions (e.g., \textit{Discus} and \textit{Ellipsoid} \cite{J_OMS_hansen2021coco}); 3) the \textit{exponentially smoothing} update exploits mainly the \textit{temporal} information but not the (parallel) \textit{spatial} information, which can be alleviated via the distributed algorithms like Meta-ES \cite{J_NACO_beyer2002evolution}.

In this subsection, we use the collective learning (or called collective intelligence by Schwefel \cite{C_Ecodyn_schwefel1988collective}) paradigm to exploit both the spatial and temporal information for CMA, since this paradigm can naturally match distributed computing. On each slave node, its corresponding ES learns different covariance matrices (directional vectors) periodically. Different form LM-CMA, for all the CMA-ES only the low-dimensional subspace is searched, which needs an extra \textit{assembling} operation in the optimizer-level of CC (see Fig. \ref{fig:hierarchical_structure}). After each relatively short learning period, all information is collected into the meta-level of DCC. Since the temporal information has been exploited on the slave nodes, we average all the covariance matrices of the better LM-CMA instances (e.g., fixed to 1/5 of all instances) as its base of the next cycle. The (weighted) averaging does not involve the expensive matrix multiplication operation at the meta-level, which otherwise can severely degrade the speedup on mainstream distributed computing systems based on the master-slave architecture (according to the well-known Amdahl’s Law). For better diversity of covariance matrix at the meta-level, we assemble a set of direction vectors from LM-CMA with covariance matrices from CMA-ES for some new LM-CMA instances in the next cycle.

Overall, it is expected to maintain the powerful invariance property of LM-CMA especially for ill-conditioned landscapes (really a challenge for CC), while keeping the fine-tuning ability of CC over a cycle of different subspaces. Our theoretical results have shown that CC tends to converge to the global optimum under certain conditions (even given any decomposition).

\subsection{Distributed Meta-ES for Global Step-Size Adaptation}

In the serial CMA-ES \cite{BS_hansen2014principled}, the global step-size adaptation is decoupled with the adaptation of covariance matrix, since they can work independently at different time scales for better convergence progress. In the standard CMA-ES form, the cumulative step-size adaptation (CSA) based on the evolution path is used to set the global step-size on-the-fly, which exploits the temporal correlation information over successive generations to speed up convergence significantly. However, in the distributed computing context, the CSA can be further improved, since the spatial information becomes accessible in parallel, resulting in the Meta-ES \cite{J_NACO_beyer2002evolution}. Following our previous work \cite{C_PPSN_duan2022collective}, we combine Meta-ES with CSA to enjoy the best of both worlds. Specifically, Meta-ES is run at the meta-level of DCC to keep spatial diversity while CSA is employed for each serial ES instance to exploit temporal correlation.

\subsection{Distribution Mean Update via Weighted Averaging}

The mean update of the normal search distribution has a significant impact on both the stability and effectiveness of DCC. Following the theoretical work from Beyer \cite{J_NACO_beyer2002evolution}, we use the (weighted) averaging strategy at the meta-level, in order to obtain the \textit{genetic repair} effect, which is also consistent with all ESs in slave nodes. Note that this simple strategy is also used in some gradient-based distributed optimizers (e.g., for DNN training). However, we have previously observed in \cite{C_PPSN_duan2022collective} that sometimes it could result in the \textit{regression} issue on some functions, which was also identified by Rudolph in \cite{C_PPSN_rudolph1990global}. To stabilize the evolution process and keep diversity at the meta-level, we also keep maintaining a proportion of \textit{elitist} LM-CMA instances in parallel at each learning period (e.g., fixed to 1/20, typically depending on the number of available CPU cores).

\subsection{Convergences and Implementations}

Since our proposed hierarchical CC framework uses the \textit{best-so-far} solution as the fitness evaluation base of subpopulations as the same as the continuous-game model, its convergence curve can be also theoretically modeled under the Main Theorem in Section \ref{main_theorem} without significant modifications. As a result, the global convergence condition of the Main Theorem can be well transferred to our new hierarchical CC framework directly.

On ill-conditioned non-separable landscapes, CC suffers easily from the much slow rate of convergence owing to strong dependencies among variables. However, on many non-separable landscapes with weak and sparse variable interactions, some modern CC could obtain very competitive and sometimes even state-of-the-art optimization performances. This is because weak and sparse interactions between subcomponents (aka well-conditioned landscapes) could approximately match the underlying assumptions behind CC. When as a general-purpose optimizer for large-scale BBO, it is highly desirable for CC to handle different fitness landscapes as many as possible. For this important aim, our proposed hierarchical CC framework combines the fine-tuning ability from CC’s decomposition (on well-conditioned landscapes) with the topology learning (aka invariance) ability of LM-CMA (in particular on ill-conditioned landscapes), in order to enjoy the best of worlds. For local optima avoidance, restart is often a simple yet efficient strategy, which can be easily integrated into our CC framework.

In summary, we use a state-of-the-art clustering computing system called \textbf{Ray} \cite{C_OSDI_moritz2018ray}, developed mainly by a research team from UC Berkeley, to implement our distributed algorithm (DCC).

\section{Numerical Experiments}

In this section, large-scale numerical experiments on a set of high-dimensional test functions are conducted, in order to show the advantages (and possible disadvantages) of DCC for large-scale black-box optimization.

\subsection{Benchmarking Algorithms}

We choose a total of 43 benchmarking baselines, which are classified into the following 10 main families: CC, ES, natural evolution strategies, cross-entropy methods, estimation of distribution algorithms, differential evolution, particle swarm optimization, genetic algorithms, simulated annealing, and random search\footnote{Here we have excluded the classical evolutionary programming, since its modern versions for continuous optimization is very similar to ES.}. For each algorithmic family, there are some standard versions and the latest large-scale variants. All their source code is taken from a recently developed open-source library for population-based optimization (called \textbf{PyPop7}\footnote{\url{https://github.com/Evolutionary-Intelligence/pypop}}), which is actively maintained now. Owing to page limits, please refer to \url{https://pypop.readthedocs.io} for their references.

\subsection{High-Dimensional Test Functions}

\begin{table}[!t]
\centering
\caption{A Set of 10 Benchmarking Functions}\label{tab:benchmark_functions}
\resizebox{0.95\textwidth}{!}{%
\begin{tabular}{c|l|l}
    \toprule[1.5pt]
    & \makecell[c]{\textbf{Function Name}}         & \makecell[c]{\textbf{Mathematic Function}} \\
    \midrule[1pt]
    & {sphere}               & $ f(x) = \sum_{i = 1}^{n} x_i^2$ \\
    & {cigar}                & $ f(x) = x_1^2 + 10 ^ 6 \sum_{i = 2}^{n} x_i^2 $ \\
    & {discus}               & $ f(x) = 10^6 x_1^2 + \sum_{i = 2}^{n} x_i^2 $ \\
    & {cigar$\_$discus}          & $ f(x) = x_1^2 + 10^4\sum_{i=2}^{n-1}x_i^2 + 10^6x_n^2 $ \\
    & {ellipsoid}            & $ f(x) = \sum_{i = 1}^{n} 10^{\frac{6(i- 1)}{n - 1}} x_i^2 $ \\
    & {different$\_$powers}      & $ f(x) = \sum_{i = 1}^{ n} \left | x_i \right | ^{\frac{2 + 4(i - 1)}{n - 1}} $ \\
    & {schwefel221}          & $ f(x) = \max(\left | x_1 \right |, \cdots, \left | x_n \right |) $ \\
    & {step}                 & $ f(x) = \sum_{i=1}^{n}(\lfloor x_i + 0.5 \rfloor)^2 $ \\
    & {rosenbrock}           & $ f(x) = 100 \sum_{i = 1}^{n -1} (x_i^2 - x_{i + 1})^2 + \sum_{i = 1}^{n - 1} (x_i - 1)^2 $ \\
    & {schwefel12}           & $ f(x) = \sum_{i=1}^{n}(\sum_{j=1}^{i}x_j)^2 $ \\
    \bottomrule[1.5pt]
\end{tabular}%
}
\end{table}%

Since the focus of this paper is \textit{non-separable} large-scale BBO, we choose 10 high-dimensional test functions, which are often used to analyze the convergence property in the ES community. For modern ESs, invariance is one of fundamental design principles when optimizing non-separable functions. See Table \ref{tab:benchmark_functions} for their detailed formula. As a standard benchmarking \cite{J_NatMI_kudela2022critical} practice, we use the \textit{rotation} and \textit{shift} operation to make all test functions non-separable and avoid the origin being the global optimum, respectively.

In this paper, we set the number of dimensions to 2000 for all test functions. Since the rotation operation\footnote{Generating rotation matrix via Schmidt orthogonalization has a cubic time complexity.} has a quadratic computational complexity, we need to efficiently utilize the shared memory of each node to avoid expensive data-transfer operations, following our previous paper \cite{C_PPSN_duan2022collective}. Note that this setting is critical for the speedup of memory-costly function evaluations for distributed EAs. For all test functions, the initial search range is set to $(-10.0, 10.0)^n$.

For fair comparisons of all 42 general-purpose black-box optimizers, we did not fine-tune any hyper-parameter settings for each optimizer on every test function independently. Instead, arguably as a standard practice, we used the same hyper-parameter values on all test functions. For all 41 baseline optimizers, we used the default values suggested from the well-designed open-source library called PyPop7, since at least these default values have been well-tested in their corresponding papers. For our own distributed algorithm, the maximum of available parallel computing units (independent CPU cores considered in our paper) is a key factor to impact the proper settings of its hyper-parameters. For simplicity, it is better to self-adapt these hyper-parameters according to the maximum of available parallel computing resources, if possible. Specifically, the total number of parallel sub-optimizers is set as the total number of available CPU (logical) cores minus the number of their corresponding (Linux) servers, in order to avoid possible program blocking owing to competition of computing resources. This is mainly because we need to leave (at least) one CPU core in each independent (Linux) server to do a series of necessary OS-level management tasks such as network communications, task scheduling, and so on.

\subsection{Experimental Setups}

On all test functions, each algorithm was totally run 5 times. We set two termination conditions for all algorithms to make fair comparisons as much as possible: 1) when the best-so-far fitness is below $1\mathrm{e}{-10}$; 2) when the maximum runtime exceeds 3 hours. As a result, the total CPU runtime \textit{roughly} equals 6450 (= 43 algorithms*10 functions*5 trails*3 hours) hours (i.e., 269 days) if they were run only on a single machine. To reduce the time-consuming benchmarking process, 10 HPC servers were used in parallel, all of which were also together used to build a (slightly heterogeneous) clustering computing platform. There are a total of 400 CPU logic cores and about 340GB memory in our private clustering computing platform. Under page limits, please see online Supplementary Materials (available at \url{https://github.com/Evolutionary-Intelligence/DCC}) for detailed hardware and software configurations.

\subsection{Experimental Results and Analyses}

\begin{figure}
    \centering
    \includegraphics[width=0.98\textwidth]{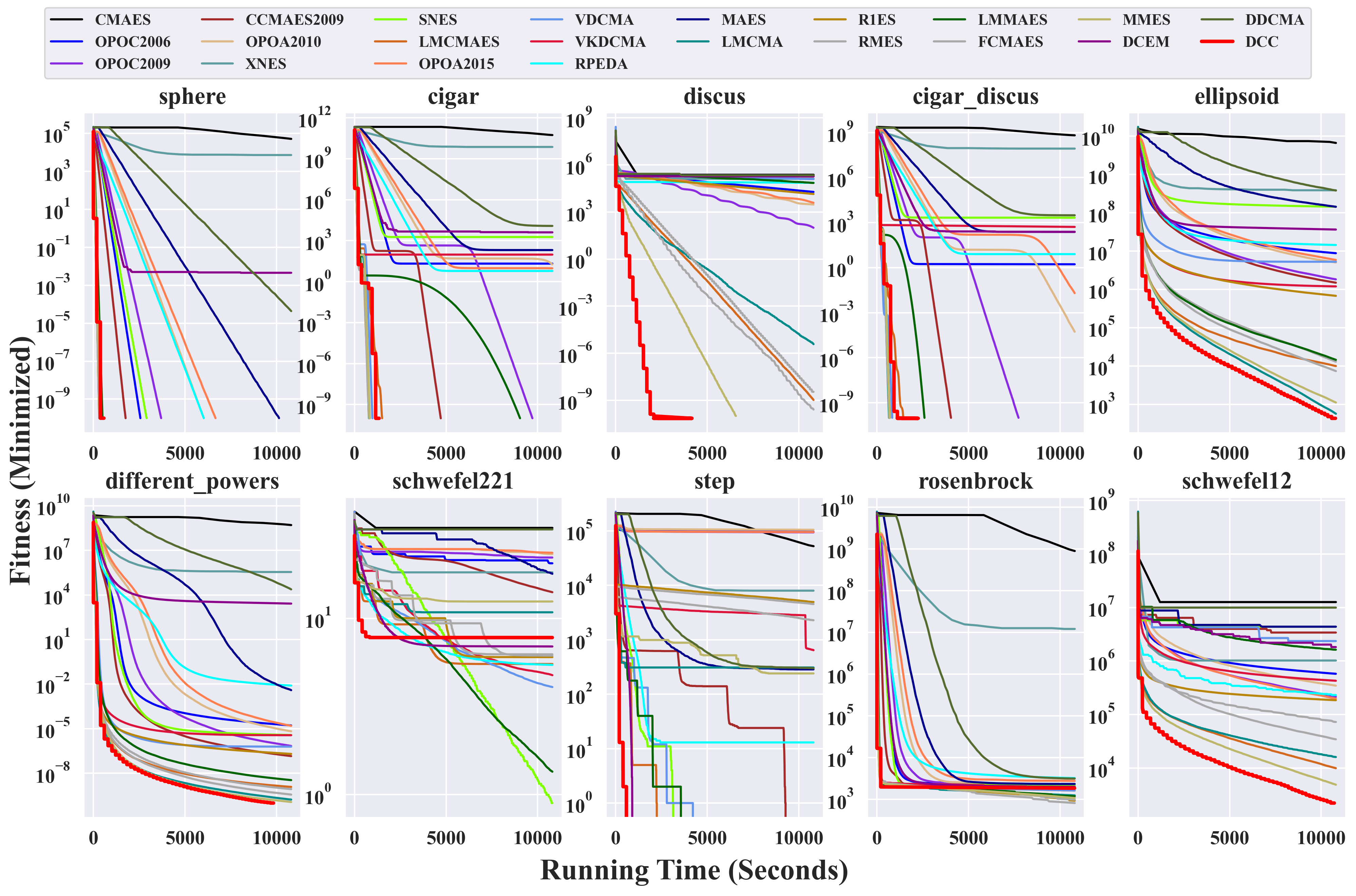}
    \caption{Median convergence curves on a set of 2000-d rotated and shifted test functions. To avoid crowd and confusion (given 44 algorithms), only the first half (22) of optimizers are drawn in each subfigure. Refer to \href{https://github.com/Evolutionary-Intelligence/DCC}{online Supplementary Materials} for detailed data.}
    \label{fig:convergence_curve_1_22}
\end{figure}


In Fig. \ref{fig:convergence_curve_1_22}, the convergence curves on all test functions are drawn for 22 black-box optimizers. To avoid crowd in plotting, all other 22 optimizers are shown in online Supplementary Materials (available at \url{https://github.com/Evolutionary-Intelligence/DCC}), since all of them are much worse than or close (only on one test function) to our distributed algorithm on these high-dimensional functions.

On five test functions (i.e., \textit{schwefel}12, \textit{discus}, \textit{ellipsoid}, \textit{different}$\_$\textit{powers}, and \textit{step}), clearly DCC showed the best convergence rates. For the first forth cases, there are multiple (rather than one predominated) promising search directions. DCC’s diversity maintaining of covariance matrix can exploit them in parallel at the meta-level. For the last case, properly setting the global step size on-the-fly is important for fast convergence rate since there exist many plateaus. For DCC, its Meta-ES-based adaptation strategy can alleviate this problem significantly, since it exploits the spatial information well (at the same time multiple parallel CSA strategies could exploit the temporal information inside each cycle).

On four other functions (i.e., \textit{cigar}, \textit{sphere}, \textit{cigar$\_$discus} and \textit{rosenbrock}), DCC still obtained very competitive results, though not the best one. For all of them except \textit{sphere} there is one predominated search direction, which can be efficiently learnt by the (exponential smoothing) evolution path. In this case, keeping the diversity on covariance matrix seems to be not the most important. The performance differences on these functions are very small (mainly from the extra overhead of distributed computing), which may be negligible. For DCC, the stabilization strategy for distribution mean update helps to maintain the advantages of the underlying suboptimizers.

However, distributed computing is \textbf{not a panacea}, which cannot escape from the No-Free-Lunch theorems \cite{J_TEVC_wolpert1997no}. On \textit{schwefel}221, DCC suffered from a slow convergence. This is due to the \textit{loss of gradient} on this \textit{min-max} function. See \ref{appendix:convergence_loss_grad} for a theoretical analysis regarding convergence.

Overall, the zig-zag-type convergence rate is reminiscent of a famous natural evolution theory named as “\textbf{punctuated equilibrium}” \cite{J_Nature_gould1993punctuated}. Such a punctuated equilibrium-style convergence curve may be an essential property of many distributed EAs, which is worthwhile to be further investigated.

\section{One Application Case on Black-Box Classification}

To validate the practical utility of our distributed framework, in this section we choose one application case from data science as the testbed, namely black-box classification. Currently, (two-class) black-box classification is commonly used to compare the performance of different zeroth-order optimizers in more realistic scenarios. Here our used dataset called QSAR androgen receptor (QAR) comes from the biological chemistry community, which is accessible at the popular UCI Machine Learning Repository. For this dataset, there are a set of 1687 instances with 1024 features.

Here we adopt a common loss function
\[
g(\mathbf{w})=\sum_{p=1}^P \max \left(0,1-y_p \stackrel{\circ}{\mathbf{x}}^T \mathbf{w}\right)
\]
where
 \(\left\{\left(\mathbf{x}_p, y_p\right)\right\}_{p=1}^P\) with the labels \(y_p \in\{-1,+1\}\) represents training dataset, \(\stackrel{\circ}{\mathbf{x}}^T \mathbf{w}=0\) is a linear decision boundary, from support vector machines (SVM). Note that there exist non-differentiable search points, since this loss function involves a non-smoothed max() operation. Therefore, this loss function has been commonly used for black-box classification. For the dataset QAR, there are 1025 (= 1024 + 1) weights to be optimized, after we consider the bias term in the SVM loss function.

\begin{figure}
    \centering
    \includegraphics[width=1.0\linewidth]{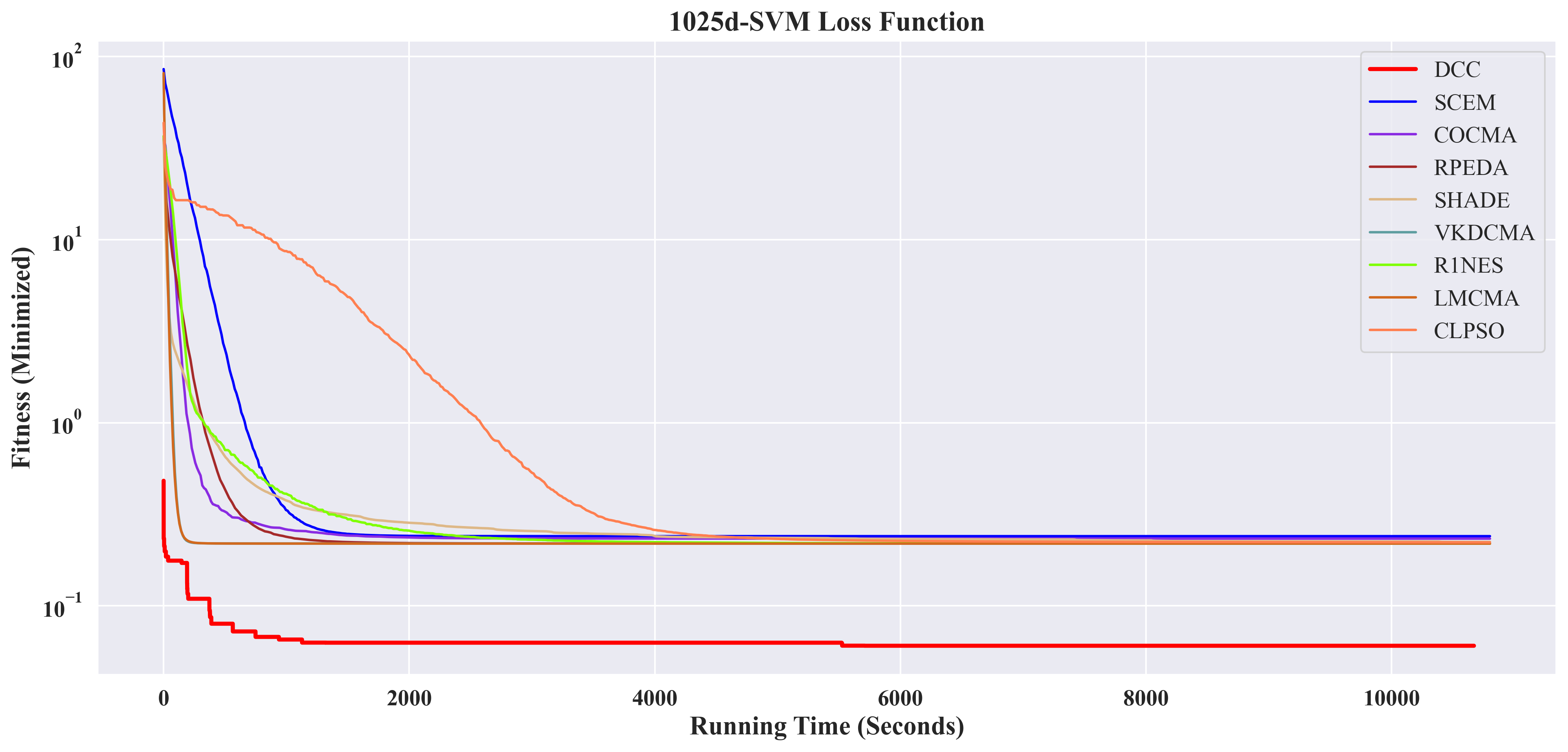}
    \caption{SVM loss function}
    \label{fig:svm-loss-function}
\end{figure}

As is shown in Fig. \ref{fig:svm-loss-function}, our distributed algorithm (DCC) can obtain the fastest rate of convergence and the best quality of solution on this high-dimensional loss function, among 9 competitive algorithms. The reason behind DCC’s superiority on this application case may lie in that our distributed algorithm could efficiently utilize these parallel computing units to collectively learn its topology/geometry structure well. In the future, we plan to test more application cases to validate the potential of DCC.

\section{Conclusion}

In this paper, we have for the first time examined the \textit{coherence} of the theoretical concept (PAS) for CC with many real-world applications, which is believed to be of central importance \cite{R_averick1992minpack2}, and showed that \textit{few} real-world applications could well match such an ideal assumption\footnote{See \url{https://tinyurl.com/3d4nnneb} for a variety of applications regarding evolutionary computation.}. Instead, we have provided a pure Nash equilibrium (PNE) perspective as a new avenue to capture the essence of CC \cite{C_GECCO_dejong2007introductory} to explain currently empirical successes on some (rather all) non-separable LSO problems. Furthermore, based on the above theoretical results, we have extended the serial version of CC to the powerful distributed computing scenario and proposed its distributed version (DCC), where collective learning \cite{J_PNAS_vanchurin2022theory,C_Ecodyn_schwefel1988collective} plays an important role when combined with two state-of-the-art ES variants (LM-CMA and CMA-ES). Numerical experiments have shown the scalability (w.r.t. CPU cores) and potentials of DCC on a set of high-dimensional benchmark functions. We plan to extend DCC in the future as follows: 1) to build a model to analyze its convergence rate; 2) to further validate its efficiency and effectiveness on some real-world applications; 3) to scale it up to more than one thousands of CPU cores (“\textit{more is different}” \cite{BS_anderson2008more}).

Furthermore, formalizing CC as a continuous game model via simplifications suffers from two main issues: 1) it does not consider other complex subcomponent interactions (e.g., lenient learners) except the widely used interaction strategy (i.e., based on only the best-so-far solution); 2) it relies heavily on an idealized assumption upon all sub-optimizers, that is, each sub-optimizer could find the optima for all subproblems under a limited (even possibly a very huge) number of function evaluations. We leave the relaxation of these two theoretical assumptions for further works.

\newpage

\appendix

\section{Convergence to a Pure Nash Equilibrium (PNE) for Cooperative Coevolution (CC)}\label{appedix:convergence_pne}

\textbf{Theorem 1}: Given \textit{any} partition $p=\left\{g_1, \ldots, g_m\right\}$ of the objective function $f(\mathbf{x}): \mathbb{R}^n \rightarrow \mathbb{R}$, where $1 <m \leq$ $n$, we say that the convergence point of CC under this partition is also one \textit{pure Nash equilibrium (PNE)}.

\textbf{Proof}: If $\mathbf{x}^*=\left(x_1^*, x_2^*, \cdots, x_n^*\right)=\left(\mathbf{x}_{g_1}^*, \cdots, \mathbf{x}_{g_m}^*\right)$ is a convergence point of CC under the partition $p=$ $\left\{g_1, \ldots, g_m\right\}$, then $\mathbf{x}_{g_i}^*$ is one of the global optima of the function $f\left(\mathbf{x}_{g_1}^*, \cdots, \mathbf{x}_{g_i}, \cdots, \mathbf{x}_{g_m}^*\right), i=1, \cdots, m$.

If $\mathbf{x}^*=\left(\mathbf{x}_{g_1}^*, \cdots, \mathbf{x}_{g_m}^*\right)$ is not one \textit{pure Nash equilibrium} (w.r.t. $p$ ), according to the definition of \textbf{Pure Nash Equilibrium (PNE)}, we have that $\exists i \in\{1, \cdots, m\}, \exists \check{\mathbf{x}}_{g_i} \in \mathbb{R}^{|g_i|} \backslash\left\{\mathbf{x}_{g_i}^*\right\}$, such that $f\left(\mathbf{x}_{g_1}^*, \cdots, \check{\mathbf{x}}_{g_i}, \cdots, \mathbf{x}_{g_m}^*\right) < f\left(\mathbf{x}_{g_1}^*, \cdots, \mathbf{x}_{g_i}^*, \cdots, \mathbf{x}_{g_m}^*\right)$, namely, then $\mathbf{x}_{g_i}^*$ is not the global optimum of $f\left(\mathbf{x}_{g_1}^*, \cdots, \mathbf{x}_{g_i}, \cdots, \mathbf{x}_{g_m}^*\right)$. It is a contradiction.

Note that we assume that for CC, the suboptimizer could obtain the global optimum for each subproblem given a \textit{limited} number of function evaluations (but which could be any large number). We admit that such an assumption appears to be difficult to satisfy in practice. However, it helps to understand the convergence behavior of CC and capture the \textit{game-theoretical} essence of CC.

\section{Convergence Analyses on Four Representative Test Functions}\label{appendix:convergence_four}

\subsection{Function 1}

$$
f_1(x, y)=7 x^2+6 x y+8 y^2
$$
Since its Hessian matrix $\left[\begin{array}{ll}7 & 6 \\ 6 & 8\end{array}\right]$ is positive definite, $f_1$ is differentiable strictly convex, then it has a unique global optimum $(0,0)$. Given $\left(x_0, y_0\right)$ is a PNE, by the definition of PNE, $\left(x_0, y_0\right)$ is the unique global optimum of differentiable strictly convex $f_1\left(x, y_0\right)$ and $f_1\left(x_0, y\right)$. Then $\frac{\partial f_1\left(x, y_0\right)}{\partial x}=14 x+6 y_0=0$, we have $x=-\frac{6 y_0}{14}=x_0, \frac{\partial f_1\left(x_0, y\right)}{\partial y}=16 y+6 x_0=0$, we have $y=$ $-\frac{6 x_0}{16}=y_0$. So, $\left(x_0, y_0\right)=(0,0)$.

\subsection{Function 2}

$$
f_2(x, y)=x^2+10^6 y^2
$$
Since its Hessian matrix $\left[\begin{array}{cc}1 & 0 \\ 0 & 10^6\end{array}\right]$ is positive definite, $f_1$ and its rotation variant are all strictly convex and have the unique global optimum $(0,0)$.
Like the above proof, they have only one PNE, namely global optimum $(0,0)$.

\subsection{Function 3}

$$
f_3(x, y)=100\left(x^2-y\right)^2+(x-1)^2
$$
Obviously, $f_3 \geq 0$ and
$$
\left\{\begin{array}{c}
\frac{\partial f_3(x, y)}{\partial x}=400 x\left(x^2-y\right)+2(x-1)=0 \\
\frac{\partial f_3(x, y)}{\partial y}=200\left(y-x^2\right)=0
\end{array},\right.
$$
at $(x, y) = (1, 1)$. So, it has a unique global optimum $(1,1)$. Suppose $\left(x_0, y_0\right)$ is a PNE, then $x_0$ and $y_0$ are the global minima of differentiable $f_3\left(x, y_0\right)$ and $f_3\left(x_0, y\right)$, respectively. Then
$$
\left\{\begin{array}{c}
\frac{d f_3\left(x, y_0\right)}{d x}=400 x^3-400 x y_0+2 x-2=0 \\
\frac{d f_3\left(x_0, y\right)}{d y}=200 y-200 x_0^2=0
\end{array},\right.
$$
we have $200 x^3-200 x x_0^2+x-1=0, y=x_0^2=y_0$.
Since $x_0$ is one of solutions of equation $200 x^3-200 x x_0^2+x-1=0$, we conclude that $x_0=1, y_0=$ 1.

\subsection{Function 4}

$$
f_4(x, y)=|x-y|-\min (x, y)=\left\{\begin{array}{c}
x-2 y, x>y \\
-x, x=y \\
y-2 x, x<y
\end{array}\right.
$$
For any $\left(x_0, y_0\right)$, the function $f_4\left(x, y_0\right)$ and $f_4\left(x_0, y\right)$ obtain their global minima at $\left(y_0, y_0\right)$ and $\left(x_0, x_0\right)$, respectively. So, the set $\{(x, y) \mid x=y\}$ is the set of PNEs.

\section{Convergence Analysis on A Function with Loss of Gradients}\label{appendix:convergence_loss_grad}

\textbf{Corollary 3}: For the \textit{Schwefel's Problem 2.21}, $f(\mathbf{x}) = \max _{i=1, \ldots, n} \left( \left| x_{i} \right| \right)$, defined on an open set\footnote{Here it is \textit{implicitly} assumed that there is (at least) one global optimum in this open set.} $\Omega \subseteq \mathbb{R}^n$, the set of global pure Nash equilibria w.r.t. any partition $p=\left\{g_1, \ldots, g_m\right\}$ is $\left\{\mathbf{x}, \max _{j \in g_1}\left|x_j\right|=\cdots=\right.$ $\left.\max _{j \in g_m}\left|x_j\right|\right\}$.
There is a unique strict global Nash equilibrium $\mathbf{x}=(0, \ldots, 0)$ w.r.t. any partition set, which equals the global optimum, and vice versa.

\textbf{Proof}: Given $\mathbf{x}=\left(\mathbf{x}_{g_1}, \cdots, \mathbf{x}_{g_m}\right)$ is a PNE w.r.t. any partition $p=\left\{g_1, \ldots, g_m\right\}$ of $f(\mathbf{x})=\max _{i=1, \ldots n}\left(\left|x_{i}\right|\right)$, owing to the definition of PNE, we have $\mathbf{x}_{g_i} \in \mathbb{R}^{\left|g_i\right|}$ for each $g_i \in p, \forall i \in\{1, \ldots, m\}$, satisfies

$$
f\left(\mathbf{x}_{g_i}, \mathbf{x}_{\neq g_i}\right) \leq f\left(\mathbf{x}_{g_i}^{\sim}, \mathbf{x}_{\neq g_i}\right), \forall \mathbf{x}_{g_i}^{\sim} \in \mathbb{R}^{\left|g_i\right|} \backslash\left\{\mathbf{x}_{g_i}\right\},
$$
namely, for any $i \in\{1, \cdots, m\}, \mathbf{x}_{g_i}^{\sim} \in \mathbb{R}^{|g_i|} \backslash \left\{\mathbf{x}_{g_i}\right\}, \max \left\{\max _{j \in g_i}\left|x_j\right|, \max _{j \in \neq g_i}\left|x_j\right|\right\} \leq \max \left\{\max _{j \in g_i}\left|x_{g_i}^{\sim}\right|\right.$, $\left.\max _{j \in \neq g_i}\left|x_j\right|\right\}$, only if, $\max \left\{\max _{j \in g_i}\left|x_j\right|, \max _{j \in \neq g_i} \left|x_j\right|\right\} \leq \min \left\{\max _{j \in g_i}\left|x_{g_i}^{\sim}\right|, \max _{j \in \neq g_i}\left|x_j\right|\right\}$, only if,
$$
\left\{\begin{array}{c}
\max _{j \in g_i}\left|x_j\right| \leq \min \left\{\max _{j \in g_i}\left|x_{g_i}^{\sim}\right|\right\} \\
\max _{j \in g_i}\left|x_j\right| \leq \max _{j \in \neq g_i}\left|x_j\right| \\
\max _{j \in \neq g_i}\left|x_j\right| \leq \min \left\{\max _{j \in g_i}\left|x_{g_i}^{\sim}\right|\right\}
\end{array}\right.
$$
owing to the definition of PNE,
$$
\max _{j \in g_i}\left|x_j\right| \leq \min \left\{\max _{j \in g_i}\left|x_{g_i}^{\sim}\right|\right\}
$$
so, we only need for any $i \in\{1, \cdots, m\}$,
$$
\left\{\begin{array}{c}
\max _{j \in g_i}\left|x_j\right| \leq \max _{j \in \neq g_i}\left|x_j\right| \\
\max _{j \in \neq g_i}\left|x_j\right| \leq \max _{j \in g_i}\left|x_j\right|
\end{array}\right.
$$
we have $\max _{j \in g_1}\left|x_j\right|=\cdots=\max _{j \in g_m}\left|x_j\right|$.

Since $f(\mathbf{x}) \geq 0, \mathbf{x_0}=(0, \cdots, 0)$ is a PNE, and $f\left(\mathbf{x_0}\right)=f(0, \cdots, 0)=\mathbf{0},  \mathbf{x_0}$ is the unique global optimum, thus it is a unique strict global Nash equilibrium, and vice versa.

\newpage

\bibliographystyle{elsarticle-num}
\bibliography{main}

\providecommand{\noopsort}[1]{}
\begin{thebibliography}{100}
\expandafter\ifx\csname url\endcsname\relax
  \def\url#1{\texttt{#1}}\fi
\expandafter\ifx\csname urlprefix\endcsname\relax\def\urlprefix{URL }\fi
\expandafter\ifx\csname href\endcsname\relax
  \def\href#1#2{#2} \def\path#1{#1}\fi

\bibitem{J_Nature_lecun2015deep}
Y.~LeCun, Y.~Bengio, G.~Hinton, \href{https://doi.org/10.1038/nature14539}{Deep
  learning}, Nature 521~(7553) (2015) 436--444.

\bibitem{J_NN_schmidhuber2015deep}
J.~Schmidhuber, \href{https://doi.org/10.1016/j.neunet.2014.09.003}{Deep
  learning in neural networks: An overview}, Neural Netw. 61 (2015) 85--117.

\bibitem{arXiv_zador2022nextgeneration}
A.~Zador, B.~Richards, et~al., \href{https://arxiv.org/abs/2210.08340}{Toward
  next-generation artificial intelligence: catalyzing the NeuroAI revolution}
  (2022).

\bibitem{J_FoCM_nesterov2017random}
Y.~Nesterov, V.~Spokoiny,
  \href{https://doi.org/10.1007/s10208-015-9296-2}{Random gradient-free
  minimization of convex functions}, Found. Comput. Math. 17~(2) (2017)
  527--566.

\bibitem{J_OMS_hansen2021coco}
N.~Hansen, A.~Auger, et~al.,
  \href{https://doi.org/10.1080/10556788.2020.1808977}{COCO: A platform for
  comparing continuous optimizers in a black-box setting}, Optim. Methods
  Softw. 36~(1) (2021) 114--144.

\bibitem{arXiv_salimans2017evolution}
T.~Salimans, J.~Ho, et~al.,
  \href{https://doi.org/10.48550/arXiv.1703.03864}{Evolution strategies as a
  scalable alternative to reinforcement learning} (2017).

\bibitem{J_Nature_mnih2015humanlevel}
V.~Mnih, K.~Kavukcuoglu, et~al.,
  \href{https://doi.org/10.1038/nature14236}{Human-level control through deep
  reinforcement learning}, Nature 518~(7540) (2015) 529--533.

\bibitem{J_Science_fan2020highresolution}
J.-x. Fan, S.-z. Shen, et~al., \href{https://doi.org/10.1126/science.aax4953}{A
  high-resolution summary of cambrian to early triassic marine invertebrate
  biodiversity}, Science 367~(6475) (2020) 272--277.

\bibitem{arXiv_jaderberg2017population}
M.~Jaderberg, V.~Dalibard, et~al.,
  \href{https://arxiv.org/abs/1711.09846}{Population based training of neural
  networks} (2017).

\bibitem{J_Science_jaderberg2019humanlevel}
M.~Jaderberg, W.~M. Czarnecki, et~al.,
  \href{https://doi.org/10.1126/science.aau6249}{Human-level performance in 3D
  multiplayer games with population-based reinforcement learning}, Science 364
  (2019) 859--865.

\bibitem{D_How}
\href{https://www.deepmind.com/blog/how-evolutionary-selection-can-train-more-capable-self-driving-cars}{How
  evolutionary selection can train more capable self driving cars}, (last
  visit: August 2, 2024).

\bibitem{J_Science_forrest1993genetic}
S.~Forrest, \href{https://doi.org/10.1126/science.8346439}{Genetic algorithms:
  Principles of natural selection applied to computation}, Science 261~(5123)
  (1993) 872--878.

\bibitem{J_Nature_eiben2015evolutionary}
A.~E. Eiben, J.~Smith, \href{https://doi.org/10.1038/nature14544}{From
  evolutionary computation to the evolution of things}, Nature 521~(7553)
  (2015) 476--482.

\bibitem{J_NatMI_miikkulainen2021biological}
R.~Miikkulainen, S.~Forrest,
  \href{https://doi.org/10.1038/s42256-020-00278-8}{A biological perspective on
  evolutionary computation}, Nat. Mach. Intell. 3~(1) (2021) 9--15.

\bibitem{C_PPSN_varelas2018comparative}
K.~Varelas, A.~Auger, et~al.,
  \href{https://doi.org/10.1007/978-3-319-99253-2\_1}{A comparative study of
  large-scale variants of CMA-ES}, in: PPSN, 2018, pp. 3--15.

\bibitem{J_TEVC_omidvar2022reviewa}
M.~N. Omidvar, X.~Li, X.~Yao,
  \href{https://doi.org/10.1109/tevc.2021.3130835}{A review of population-based
  metaheuristics for large-scale black-box global optimization\textemdash Part
  I}, IEEE Trans. Evol. Comput. 26~(5) (2022) 802--822.

\bibitem{J_TEVC_omidvar2022review}
M.~N. Omidvar, X.~Li, X.~Yao,
  \href{https://doi.org/10.1109/TEVC.2021.3130835}{A review of population-based
  metaheuristics for large-scale black-box global optimization\textemdash Part
  II}, IEEE Trans. Evol. Comput. 26~(5) (2022) 823--843.

\bibitem{B_nesterov2018lectures}
Y.~Nesterov, Lectures on Convex Optimization, 2nd Edition, Springer, 2018.

\bibitem{C_PPSN_potter1994cooperative}
M.~A. Potter, K.~A. Jong, \href{https://doi.org/10.1007/3-540-58484-6\_269}{A
  cooperative coevolutionary approach to function optimization}, in: PPSN,
  1994, pp. 249--257.

\bibitem{B_potter1997design}
M.~A. Potter, \href{https://cs.gmu.edu/~mpotter/pubs/thesis2.pdf}{The Design
  and Analysis of a Computational Model of Cooperative Coevolution}, Ph.D.
  thesis, George Mason University, Fairfax, VAUnited States (1997).

\bibitem{J_ECJ_potter2000cooperative}
M.~A. Potter, K.~A.~D. Jong,
  \href{https://doi.org/10.1162/106365600568086}{Cooperative coevolution: An
  architecture for evolving coadapted subcomponents}, Evol. Comput. 8~(1)
  (2000) 1--29.

\bibitem{J_SMO_gandomi2023variable}
A.~H. Gandomi, K.~Deb, et~al.,
  \href{https://doi.org/10.1007/s00158-022-03435-2}{Variable functioning and
  its application to large scale steel frame design optimization}, Struct.
  Multidisc. Optim. 66~(1) (2023) 13.

\bibitem{J_InfSci_yang2008large}
Z.~Yang, K.~Tang, X.~Yao,
  \href{https://doi.org/10.1016/j.ins.2008.02.017}{Large scale evolutionary
  optimization using cooperative coevolution}, Inf. Sci. 178~(15) (2008)
  2985--2999.

\bibitem{J_TEVC_ma2019survey}
X.~Ma, X.~Li, et~al., \href{https://doi.org/10.1109/tevc.2018.2868770}{A survey
  on cooperative co-evolutionary algorithms}, IEEE Trans. Evol. Comput. 23~(3)
  (2019) 421--441.

\bibitem{J_ECJ_hansen2003reducing}
N.~Hansen, S.~D. M{\"u}ller, P.~Koumoutsakos,
  \href{https://doi.org/10.1162/106365603321828970}{Reducing the time
  complexity of the derandomized evolution strategy with covariance matrix
  adaptation (CMA-ES)}, Evol. Comput. 11~(1) (2003) 1--18.

\bibitem{J_ECJ_akimoto2020diagonal}
Y.~Akimoto, N.~Hansen, \href{https://doi.org/10.1162/evco\_a\_00260}{Diagonal
  acceleration for covariance matrix adaptation evolution strategies}, Evol.
  Comput. 28~(3) (2020) 405--435.

\bibitem{J_JMLR_ollivier2017informationgeometric}
Y.~Ollivier, L.~Arnold, et~al.,
  \href{https://jmlr.org/papers/v18/14-467.html}{Information-geometric
  optimization algorithms: A unifying picture via invariance principles}, J.
  Mach. Learn. Res. 18~(18) (2017) 1--65.

\bibitem{BS_hansen2014principled}
N.~Hansen, A.~Auger,
  \href{http://doi.org/10.1007/978-3-642-33206-7_8}{Principled design of
  continuous stochastic search: From theory to practice}, in: Theory and
  principled methods for the design of metaheuristics, Springer, 2014, pp.
  145--180.

\bibitem{J_NACO_beyer2002evolution}
H.-G. Beyer, H.-P. Schwefel,
  \href{https://doi.org/10.1023/A:1015059928466}{Evolution strategies
  \textendash{} A comprehensive introduction}, Nat. Comput. 1 (2002) 3--52.

\bibitem{J_JMLR_wierstra2014natural}
D.~Wierstra, T.~Schaul, et~al.,
  \href{http://jmlr.org/papers/v15/wierstra14a.html}{Natural evolution
  strategies}, J. Mach. Learn. Res. 15~(27) (2014) 949--980.

\bibitem{J_Algo_akimoto2012theoretical}
Y.~Akimoto, Y.~Nagata, et~al.,
  \href{https://doi.org/10.1007/s00453-011-9564-8}{Theoretical foundation for
  CMA-ES from information geometry perspective}, Algorithmica 64~(4) (2012)
  698--716.

\bibitem{J_TEVC_omidvar2014cooperative}
M.~N. Omidvar, X.~Li, et~al.,
  \href{https://doi.org/10.1109/tevc.2013.2281543}{Cooperative co-evolution
  with differential grouping for large scale optimization}, IEEE Trans. Evol.
  Comput. 18~(3) (2014) 378--393.

\bibitem{J_ECJ_muhlenbein1999fdaa}
H.~M{\"u}hlenbein, T.~Mahnig,
  \href{https://doi.org/10.1162/evco.1999.7.4.353}{FDA-A scalable evolutionary
  algorithm for the optimization of additively decomposed functions}, Evol.
  Comput. 7~(4) (1999) 353--376.

\bibitem{R_tang2009benchmark}
K.~Tang, X.~Li, et~al., Benchmark functions for the cec'2010 special session
  and competition on large-scale global optimization, Tech. rep., Nature
  Inspired Computation and Applications Laboratory, USTC, China (2009).

\bibitem{R_li2013benchmark}
X.~Li, K.~Tang, et~al.,
  \href{https://www.tflsgo.org/assets/cec2018/cec2013-lsgo-benchmark-tech-report.pdf}{Benchmark
  functions for the CEC'2013 special session and competition on large-scale
  global optimization}, Tech. rep., Evolutionary Computing and Machine Learning
  (ECML), School of Computer Science and Information Technology, RMIT
  University, Melbourne, Australia (2013).

\bibitem{J_InfSci_omidvar2015designing}
M.~N. Omidvar, X.~Li, K.~Tang,
  \href{https://doi.org/10.1016/j.ins.2014.12.062}{Designing benchmark problems
  for large-scale continuous optimization}, Inf. Sci. 316 (2015) 419--436.

\bibitem{J_Nature_bonabeau2000inspiration}
E.~Bonabeau, M.~Dorigo, G.~Theraulaz,
  \href{https://doi.org/10.1038/35017500}{Inspiration for optimization from
  social insect behaviour}, Nature 406~(6791) (2000) 39--42.

\bibitem{J_TMTT_feng2016parallel}
F.~Feng, C.~Zhang, et~al.,
  \href{https://doi.org/10.1109/tmtt.2016.2605666}{Parallel decomposition
  approach to gradient-based EM optimization}, IEEE Trans. Microw. Theory
  Techn. 64~(11) (2016) 3380--3399.

\bibitem{J_RSIF_cheney2018scalable}
N.~Cheney, J.~Bongard, et~al.,
  \href{https://doi.org/10.1098/rsif.2017.0937}{Scalable co-optimization of
  morphology and control in embodied machines}, J. R. Soc. Interface 15~(143)
  (2018) 20170937.

\bibitem{J_TEVC_farahmand2010interaction}
A.-m. Farahmand, M.~N. Ahmadabadi, et~al.,
  \href{https://doi.org/10.1109/tevc.2009.2016216}{Interaction of culture-based
  learning and cooperative co-evolution and its application to automatic
  behavior-based system design}, IEEE Trans. Evol. Comput. 14~(1) (2010)
  23--57.

\bibitem{C_PPSN_vidal2010threshold}
F.~P. Vidal, E.~Lutton, et~al.,
  \href{https://doi.org/10.1007/978-3-642-15844-5\_42}{Threshold selection,
  mitosis and dual mutation in cooperative co-evolution: Application to medical
  3D tomography}, in: PPSN, 2010, pp. 414--423.

\bibitem{C_GECCO_rainville2013sustainable}
F.-M.~D. Rainville, M.~Sebag, et~al.,
  \href{https://doi.org/10.1145/2463372.2463556}{Sustainable cooperative
  coevolution with a multi-armed bandit}, in: GECCO, 2013, pp. 1517--1524.

\bibitem{J_TPAMI_zhai2016making}
Y.~Zhai, Y.-S. Ong, I.~W. Tsang,
  \href{https://doi.org/10.1109/tpami.2016.2533384}{Making trillion
  correlations feasible in feature grouping and selection}, IEEE Trans. Pattern
  Anal. Mach. Intell. 38~(12) (2016) 2472--2486.

\bibitem{J_TEVC_he2016cooperative}
S.~He, G.~Jia, et~al.,
  \href{https://doi.org/10.1109/tevc.2016.2530311}{Cooperative co-evolutionary
  module identification with application to cancer disease module discovery},
  IEEE Trans. Evol. Comput. 20~(6) (2016) 874--891.

\bibitem{J_TNNLS_fan2017collective}
J.~Fan, J.~Wang, \href{https://doi.org/10.1109/tnnls.2016.2582381}{A collective
  neurodynamic optimization approach to nonnegative matrix factorization}, IEEE
  Trans. Neural Netw. Learn. Syst. 28~(10) (2017) 2344--2356.

\bibitem{J_TNNLS_gong2021evolving}
M.~Gong, J.~Liu, et~al.,
  \href{https://doi.org/10.1109/tnnls.2020.2978857}{Evolving deep neural
  networks via cooperative coevolution with backpropagation}, IEEE Trans.
  Neural Netw. Learn. Syst. 32~(1) (2021) 420--434.

\bibitem{J_TMIS_rashid2022anomaly}
A.~B. Rashid, M.~Ahmed, et~al., \href{https://doi.org/10.1145/3495165}{Anomaly
  detection in cybersecurity datasets via cooperative co-evolution-based
  feature selection}, ACM Trans. Mag. Inf. Sys. 13~(3) (2022) 1--39.

\bibitem{J_TEVC_liu2022surrogateassisted}
S.~Liu, H.~Wang, et~al., \href{https://doi.org/10.1109/tevc.2022.3149601}{A
  surrogate-assisted evolutionary feature selection algorithm with parallel
  random grouping for high-dimensional classification}, IEEE Trans. Evol.
  Comput. 26~(5) (2022) 1087--1101.

\bibitem{J_TCYB_zhao2020evolutionary}
T.-F. Zhao, W.-N. Chen, et~al.,
  \href{https://doi.org/10.1109/tcyb.2020.2975530}{Evolutionary
  divide-and-conquer algorithm for virus spreading control over networks}, IEEE
  Trans. Cybern. 51~(7) (2020) 3752--3766.

\bibitem{B_james2021introduction}
G.~James, D.~Witten, et~al.,
  \href{https://doi.org/10.1007/978-1-0716-1418-1}{An Introduction to
  Statistical Learning: with Applications in R}, 2nd Edition, Springer, 2021.

\bibitem{J_JMLR_gomez2008accelerated}
F.~J. Gomez, J.~Schmidhuber, R.~Miikkulainen,
  \href{https://dl.acm.org/citation.cfm?id=1390712}{Accelerated neural
  evolution through cooperatively coevolved synapses}, J. Mach. Learn. Res. 9
  (2008) 937--965.

\bibitem{D_EvoTorch}
\href{https://evotorch.ai}{evotorch.ai}, (last visit: August 2, 2024).

\bibitem{J_NECO_schmidhuber2007training}
J.~Schmidhuber, D.~Wierstra, et~al.,
  \href{https://doi.org/10.1162/neco.2007.19.3.757}{Training recurrent networks
  by evolino}, Neural Comput. 19~(3) (2007) 757--779.

\bibitem{C_GECCO_gomez2005coevolving}
F.~J. Gomez, J.~Schmidhuber,
  \href{https://doi.org/10.1145/1068009.1068092}{Co-evolving recurrent neurons
  learn deep memory POMDPs}, in: GECCO, 2005, pp. 491--498.

\bibitem{C_ICML_fan2003utilizing}
J.~Fan, R.~Lau, R.~Miikkulainen,
  \href{https://aaai.org/Library/ICML/2003/icml03-025.php}{Utilizing domain
  knowledge in neuroevolution}, in: ICML, 2003, pp. 170--177.

\bibitem{C_IJCAI_gomez1999solvinga}
F.~J. Gomez, R.~Miikkulainen, et~al.,
  \href{https://ijcai.org/Proceedings/99-2/Papers/097.pdf}{Solving
  non-Markovian control tasks with neuroevolution}, in: IJCAI, 1999, pp.
  1356--1361.

\bibitem{J_ML_moriarty1996efficient}
D.~E. Moriarty, R.~Mikkulainen,
  \href{https://doi.org/10.1023/A:1018004120707}{Efficient Reinforcement
  Learning through Symbiotic Evolution}, Mach. Learn. 22~(1/2/3) (1996) 11--32.

\bibitem{C_ICML_moriarty1995efficient}
D.~E. Moriarty, R.~Miikkulainen,
  \href{https://doi.org/10.1016/B978-1-55860-377-6.50056-6}{Efficient learning
  from delayed rewards through symbiotic evolution}, in: ICML, 1995, pp.
  396--404.

\bibitem{J_MP_wright2015coordinate}
S.~J. Wright, \href{https://doi.org/10.1007/s10107-015-0892-3}{Coordinate
  descent algorithms}, Math. Program. 151~(1) (2015) 3--34.

\bibitem{J_ECJ_loshchilov2017lmcma}
I.~Loshchilov, \href{https://doi.org/10.1162/EVCO\_a\_00168}{LM-CMA: An
  alternative to L-BFGS for large-scale black box optimization}, Evol. Comput.
  25~(1) (2017) 143--171.

\bibitem{J_MP_liu1989limited}
D.~C. Liu, J.~Nocedal, \href{https://doi.org/10.1007/bf01589116}{On the limited
  memory BFGS method for large scale optimization}, Math. Program. 45~(1-3)
  (1989) 503--528.

\bibitem{C_CEC_sun2019decomposition}
Y.~Sun, X.~Li, et~al.,
  \href{https://doi.org/10.1109/CEC.2019.8790204}{Decomposition for large-scale
  optimization problems with overlapping components}, in: CEC, 2019, pp.
  326--333.

\bibitem{BS_conn1989introduction}
A.~R. Conn, {\relax NIM}.~Gould, P.~L. Toint,
  \href{https://researchportal.unamur.be/en/publications/an-introduction-to-the-structure-of-large-scale-nonlinear-optimiz}{An
  Introduction to the Structure of Large Scale Nonlinear Optimization Problems
  and the LANCELOT Project}, in: Computing Methods in Applied Sciences and
  Engineering, SIAM, 1989, pp. 42--51.

\bibitem{J_Nature_lipson2000automatic}
H.~Lipson, J.~B. Pollack, \href{https://doi.org/10.1038/35023115}{Automatic
  design and manufacture of robotic lifeforms}, Nature 406~(6799) (2000)
  974--978.

\bibitem{C_AAMAS_panait2007theoretical}
L.~Panait, K.~Tuyls, \href{https://doi.org/10.1145/1329125.1329173}{Theoretical
  advantages of lenient q-learners: An evolutionary game theoretic perspective.
  an evolutionary game theoretic perspective}, in: AAMAS, 2007, pp. 40:1--8.

\bibitem{C_ICLR_chen2021molecule}
B.~Chen, T.~Wang, et~al.,
  \href{https://openreview.net/forum?id=jHefDGsorp5}{Molecule optimization by
  explainable evolution}, in: ICLR, 2021.

\bibitem{C_NeurIPS_greff2017neural}
K.~Greff, S.~Van~Steenkiste, J.~Schmidhuber,
  \href{https://proceedings.neurips.cc/paper/2017/hash/d2cd33e9c0236a8c2d8bd3fa91ad3acf-Abstract.html}{Neural
  expectation maximization}, in: NeurIPS, 2017.

\bibitem{J_Science_leiserson2020there}
C.~E. Leiserson, N.~C. Thompson, et~al.,
  \href{https://doi.org/10.1126/science.aam9744}{There's plenty of room at the
  Top: What will drive computer performance after Moore's law?}, Science
  368~(6495) (2020) eaam9744.

\bibitem{J_SIOPT_nesterov2012efficiency}
{\relax Yu}.~Nesterov, \href{https://doi.org/10.1137/100802001}{Efficiency of
  coordinate descent methods on huge-scale optimization problems}, SIAM J.
  Optim. 22~(2) (2012) 341--362.

\bibitem{J_JMLR_richtarik2016distributed}
P.~Richt{\'a}rik, M.~Tak{\'a}{\v c},
  \href{https://www.jmlr.org/papers/v17/15-001.html}{Distributed coordinate
  descent method for learning with big data}, The Journal of Machine Learning
  Research 17~(75) (2016) 1--25.

\bibitem{arXiv_shi2017primer}
H.-J.~M. Shi, S.~Tu, et~al., \href{http://arxiv.org/abs/1610.00040}{A primer on
  coordinate descent algorithms} (2017).

\bibitem{J_TAC_ratliff2016characterization}
L.~J. Ratliff, S.~A. Burden, S.~S. Sastry,
  \href{https://doi.org/10.1109/tac.2016.2583518}{On the characterization of
  local Nash equilibria in continuous games}, IEEE Trans. Autom. Control 61~(8)
  (2016) 2301--2307.

\bibitem{J_ECJ_panait2010theoretical}
L.~Panait, \href{https://doi.org/10.1162/evco\_a\_00004}{Theoretical
  convergence guarantees for cooperative coevolutionary algorithms}, Evol.
  Comput. 18~(4) (2010) 581--615.

\bibitem{B_wiegand2004analysis}
R.~P. Wiegand, \href{https://dl.acm.org/doi/10.5555/997339}{An Analysis of
  Cooperative Coevolutionary Algorithms}, Ph.D. thesis, George Mason
  University, Fairfax, VAUnited States (2004).

\bibitem{C_PPSN_wiegand2004spatial}
R.~P. Wiegand, J.~Sarma,
  \href{https://doi.org/10.1007/978-3-540-30217-9\_92}{Spatial embedding and
  loss of gradient in cooperative coevolutionary algorithms}, in: PPSN, 2004,
  pp. 912--921.

\bibitem{C_PPSN_duan2022collective}
Q.~Duan, G.~Zhou, et~al.,
  \href{https://doi.org/10.1007/978-3-031-14721-0\_20}{Collective learning of
  low-memory matrix adaptation for large-scale black-box optimization}, in:
  PPSN, 2022, pp. 281--294.

\bibitem{J_PNAS_vanchurin2022theory}
V.~Vanchurin, Y.~I. Wolf, et~al.,
  \href{https://doi.org/10.1073/pnas.2120037119}{Toward a theory of evolution
  as multilevel learning}, Proc. Natl. Acad. Sci. USA 119~(6) (2022)
  e2120037119.

\bibitem{J_ASOC_hansen2011impacts}
N.~Hansen, R.~Ros, et~al.,
  \href{https://doi.org/10.1016/j.asoc.2011.03.001}{Impacts of invariance in
  search: When CMA-ES and PSO face ill-conditioned and non-separable problems},
  Appl. Soft. Comput. 11~(8) (2011) 5755--5769.

\bibitem{J_TEVC_vandenbergh2004cooperative}
F.~{\noopsort{bergh}}{van den Bergh}, A.~P. Engelbrecht,
  \href{https://doi.org/10.1109/TEVC.2004.826069}{A Cooperative approach to
  particle swarm optimization}, IEEE Trans. Evol. Comput. 8~(3) (2004)
  225--239.

\bibitem{C_PPSN_chen2010largescale}
W.~Chen, T.~Weise, et~al.,
  \href{https://doi.org/10.1007/978-3-642-15871-1\_31}{Large-scale global
  optimization using cooperative coevolution with variable interaction
  learning}, in: PPSN, 2010, pp. 300--309.

\bibitem{J_TEVC_li2012cooperatively}
X.~Li, X.~Yao, \href{https://doi.org/10.1109/TEVC.2011.2112662}{Cooperatively
  coevolving particle swarms for large scale optimization}, IEEE Trans. Evol.
  Comput. 16~(2) (2012) 210--224.

\bibitem{J_TOMS_mei2016competitive}
Y.~Mei, M.~N. Omidvar, et~al., \href{https://doi.org/10.1145/2791291}{A
  competitive divide-and-conquer algorithm for unconstrained large-scale
  black-box optimization}, ACM Trans. Math. Softw. 42~(2) (2016) 13:1--24.

\bibitem{J_TEVC_sabar2017heterogeneous}
N.~R. Sabar, J.~Abawajy, J.~Yearwood,
  \href{https://doi.org/10.1109/tevc.2016.2602860}{Heterogeneous cooperative
  co-evolution memetic differential evolution algorithm for big data
  optimization problems}, IEEE Trans. Evol. Comput. 21~(2) (2017) 315--327.

\bibitem{J_TEVC_yang2017efficient}
M.~Yang, M.~N. Omidvar, et~al.,
  \href{https://doi.org/10.1109/tevc.2016.2627581}{Efficient resource
  allocation in cooperative co-evolution for large-scale global optimization},
  IEEE Trans. Evol. Comput. 21~(4) (2017) 493--505.

\bibitem{J_TEVC_omidvar2017dg2}
M.~N. Omidvar, M.~Yang, et~al.,
  \href{https://doi.org/10.1109/tevc.2017.2694221}{DG2: A faster and more
  accurate differential grouping for large-scale black-box optimization}, IEEE
  Trans. Evol. Comput. 21~(6) (2017) 929--942.

\bibitem{J_TCYB_ge2017cooperative}
H.~Ge, L.~Sun, et~al.,
  \href{https://doi.org/10.1109/tcyb.2017.2685944}{Cooperative hierarchical PSO
  with two stage variable interaction reconstruction for large scale
  optimization}, IEEE Trans. Cybern. 47~(9) (2017) 2809--2823.

\bibitem{J_TEVC_sun2018recursive}
Y.~Sun, M.~Kirley, S.~K. Halgamuge,
  \href{https://doi.org/10.1109/TEVC.2017.2778089}{A recursive decomposition
  method for large scale continuous optimization}, IEEE Trans. Evol. Comput.
  22~(5) (2018) 647--661.

\bibitem{J_TCYB_ren2018boosting}
Z.~Ren, Y.~Liang, et~al.,
  \href{https://doi.org/10.1109/tcyb.2018.2859635}{Boosting cooperative
  coevolution for large scale optimization with a fine-grained computation
  resource allocation strategy}, IEEE Trans. Cybern. 49~(12) (2018) 4180--4193.

\bibitem{J_ECJ_wang2018cooperative}
Y.~Wang, H.~Liu, et~al.,
  \href{https://doi.org/10.1162/evco\_a\_00214}{Cooperative coevolution with
  formula-based variable grouping for large-scale global optimization}, Evol.
  Comput. 26~(4) (2018) 569--596.

\bibitem{J_TCYB_peng2018multimodal}
X.~Peng, Y.~Jin, H.~Wang,
  \href{https://doi.org/10.1109/tcyb.2018.2846179}{Multimodal optimization
  enhanced cooperative coevolution for large-scale optimization}, IEEE Trans.
  Cybern. 49~(9) (2018) 3507--3520.

\bibitem{J_TEVC_yang2021efficient}
M.~Yang, A.~Zhou, et~al., \href{https://doi.org/10.1109/tevc.2020.3009390}{An
  efficient recursive differential grouping for large-scale continuous
  problems}, IEEE Trans. Evol. Comput. 25~(1) (2021) 159--171.

\bibitem{J_TEVC_liu2020hybrid}
H.~Liu, Y.~Wang, N.~Fan, \href{https://doi.org/10.1109/tevc.2020.2985672}{A
  hybrid deep grouping algorithm for large scale global optimization}, IEEE
  Trans. Evol. Comput. 24~(6) (2020) 1112--1124.

\bibitem{J_TELO_xu2021constraintobjective}
P.~Xu, W.~Luo, et~al.,
  \href{https://doi.org/10.1145/3469036}{Constraint-objective cooperative
  coevolution for large-scale constrained optimization}, ACM Trans. Evol.
  Learn. Optim. 1~(3) (2021) 1--26.

\bibitem{J_TEVC_chen2023efficient}
A.~Chen, Z.~Ren, et~al., \href{https://doi.org/10.1109/tevc.2022.3170793}{An
  efficient adaptive differential grouping algorithm for large-scale black-box
  optimization}, IEEE Trans. Evol. Comput. 27~(3) (2023) 475--489.

\bibitem{J_TEVC_ma2022merged}
X.~Ma, Z.~Huang, et~al.,
  \href{https://doi.org/10.1109/tevc.2022.3144684}{Merged differential grouping
  for large-scale global optimization}, IEEE Trans. Evol. Comput. 26~(6) (2022)
  1439--1451.

\bibitem{J_TEVC_wu2023cooperative}
Y.~Wu, X.~Peng, et~al.,
  \href{https://doi.org/10.1109/tevc.2022.3180224}{Cooperative coevolutionary
  CMA-ES with landscape-aware grouping in noisy environments}, IEEE Trans.
  Evol. Comput. 27~(3) (2023) 686--700.

\bibitem{J_TEVC_kumar2022efficient}
A.~Kumar, S.~Das, R.~Mallipeddi,
  \href{https://doi.org/10.1109/tevc.2022.3230070}{An efficient differential
  grouping algorithm for large-scale global optimization}, IEEE Trans. Evol.
  Comput. (2022) in press.

\bibitem{J_TEVC_xu2022difficulty}
P.~Xu, W.~Luo, et~al.,
  \href{https://doi.org/10.1109/tevc.2022.3201691}{Difficulty and contribution
  based cooperative coevolution for large-scale optimization}, IEEE Trans.
  Evol. Comput. (2022) in press.

\bibitem{C_NIPS_mitchell1993when}
M.~Mitchell, J.~Holland, S.~Forrest,
  \href{https://proceedings.neurips.cc/paper/1993/hash/ab88b15733f543179858600245108dd8-Abstract.html}{When
  will a genetic algorithm outperform hill climbing}, in: NeurIPS, 1993, pp.
  51--58.

\bibitem{J_TEVC_chen2019cooperative}
W.-N. Chen, Y.-H. Jia, et~al.,
  \href{https://doi.org/10.1109/tevc.2019.2893447}{A cooperative
  co-evolutionary approach to large-scale multisource water distribution
  network optimization}, IEEE Trans. Evol. Comput. 23~(5) (2019) 842--857.

\bibitem{J_TEVC_chen2022decomposition}
M.~Chen, W.~Du, et~al., \href{https://doi.org/10.1109/tevc.2022.3218375}{A
  decomposition method for both additively and non-additively separable
  problems}, IEEE Trans. Evol. Comput. (2022) in press.

\bibitem{J_TCYB_li2023dual}
J.-Y. Li, Z.-H. Zhan, et~al.,
  \href{https://doi.org/10.1109/tcyb.2022.3158391}{Dual differential grouping:
  A more general decomposition method for large-scale optimization}, IEEE
  Trans. Cybern. 53~(6) (2023) 3624--3638.

\bibitem{J_TEVC_komarnicki2022incremental}
M.~M. Komarnicki, M.~W. Przewozniczek, et~al.,
  \href{https://doi.org/10.1109/tevc.2022.3216968}{Incremental recursive
  ranking grouping for large scale global optimization}, IEEE Trans. Evol.
  Comput. (2022) in press.

\bibitem{J_TEVC_zhang2019dynamic}
X.-Y. Zhang, Y.-J. Gong, et~al.,
  \href{https://doi.org/10.1109/tevc.2019.2895860}{Dynamic cooperative
  coevolution for large scale optimization}, IEEE Trans. Evol. Comput. 23~(6)
  (2019) 935--948.

\bibitem{J_TCYB_jia2020contributionbased}
Y.~Jia, Y.~Mei, M.~Zhang,
  \href{https://doi.org/10.1109/TCYB.2020.3025577}{Contribution-based
  cooperative co-evolution for nonseparable large-scale problems with
  overlapping subcomponents}, IEEE Trans. Cybern. 52~(6) (2020) 4246--4259.

\bibitem{J_TSMC_zhang2023graphbased}
X.~Zhang, B.-W. Ding, et~al.,
  \href{https://doi.org/10.1109/tsmc.2022.3212045}{Graph-based deep
  decomposition for overlapping large-scale optimization problems}, IEEE Trans.
  Syst., Man, Cybern. Syst. 53~(4) (2023) 2374--2386.

\bibitem{C_FOGA_wiegand2002modeling}
R.~P. Wiegand, W.~C. Liles, K.~A. De~Jong,
  \href{http://eecs.ucf.edu/~wiegand/papers/Wiegand2002foga.pdf}{Modeling
  variation in cooperative coevolution using evolutionary game theory.}, in:
  Proceedings of the Seventh Workshop on Foundations of Genetic Algorithms,
  2002, pp. 203--220.

\bibitem{J_TEVC_ficici2005gametheoretic}
S.~G. Ficici, O.~Melnik, J.~B. Pollack,
  \href{https://doi.org/10.1109/tevc.2005.856203}{A game-theoretic and
  dynamical-systems analysis of selection methods in coevolution}, IEEE Trans.
  Evol. Comput. 9~(6) (2005) 580--602.

\bibitem{C_GECCO_jansen2003exploring}
T.~Jansen, R.~P. Wiegand,
  \href{https://doi.org/10.1007/3-540-45105-6\_37}{Exploring the explorative
  advantage of the cooperative coevolutionary (1+ 1) EA}, in: Genetic and
  Evolutionary Computation Conference, 2003, pp. 310--321.

\bibitem{J_ECJ_jansen2004cooperative}
T.~Jansen, R.~P. Wiegand, \href{https://doi.org/10.1162/1063656043138905}{The
  cooperative coevolutionary (1+ 1) EA}, Evol. Comput. 12~(4) (2004) 405--434.

\bibitem{R_toint1984test}
P.~L. Toint, Test problems for partially separable optimization and results for
  the routine pspmin, Tech. rep., Department of Mathematics, The University of
  Namur, Belgium (1984).

\bibitem{J_TOMS_porcelli2022exploiting}
M.~Porcelli, P.~L. Toint, \href{https://doi.org/10.1145/3474054}{Exploiting
  problem structure in derivative free optimization}, ACM Trans. Math. Softw.
  48~(1) (2022) 1--25.

\bibitem{R_averick1992minpack2}
B.~Averick, R.~Carter, et~al., \href{https://doi.org/10.2172/79972}{The
  MINPACK-2 test problem collection}, Tech. rep., Office of Scientific and
  Technical Information / Argonne National Laboratory Argonne, Argonne, IL, USA
  (1992).

\bibitem{J_COA_bouaricha1997impact}
A.~Bouaricha, J.~J. Mor{\`e},
  \href{https://doi.org/10.1023/a:1008628114432}{Impact of partial separability
  on large-scale optimization}, Comput. Optim. Appl. 7 (1997) 27--40.

\bibitem{J_JASA_hildreth1955corrigenda}
C.~Hildreth, \href{https://doi.org/10.2307/2281222}{Corrigenda: Point estimates
  of ordinates of concave functions}, J. Am. Stat. Assoc. 50~(272) (1955) 1331.

\bibitem{D_A000110}
\href{https://oeis.org/A000110/list}{oeis.org}, (last visit: August 2, 2024).

\bibitem{J_PNAS_nash1950equilibrium}
J.~F. Nash, \href{https://doi.org/10.1073/pnas.36.1.48}{Equilibrium points in
  {\emph{n}} -person games}, Proc. Natl. Acad. Sci. USA 36~(1) (1950) 48--49.

\bibitem{J_AnnMath_nash1951noncooperative}
J.~Nash, \href{https://doi.org/10.2307/1969529}{Non-cooperative games}, Ann.
  Math. 54~(2) (1951) 286.

\bibitem{C_CEC_duan2019when}
Q.~Duan, C.~Shao, et~al., \href{https://doi.org/10.1109/CEC.2019.8790148}{When
  cooperative co-evolution meets coordinate descent: Theoretically deeper
  understandings and practically better implementations}, in: CEC, 2019, pp.
  721--730.

\bibitem{J_ECJ_shang2006note}
Y.-W. Shang, Y.-H. Qiu, \href{https://doi.org/10.1162/evco.2006.14.1.119}{A
  note on the extended Rosenbrock function}, Evol. Comput. 14~(1) (2006)
  119--126.

\bibitem{J_JSIAM_warga1963minimizing}
J.~Warga, \href{https://doi.org/10.1137/0111043}{Minimizing certain convex
  functions}, J. Soc. Ind. Appl. Math. 11~(3) (1963) 588--593.

\bibitem{J_ECMA_rosen1965existence}
J.~B. Rosen, \href{https://doi.org/10.2307/1911749}{Existence and uniqueness of
  equilibrium points for concave n-person games}, Econometrica 33~(3) (1965)
  520--534.

\bibitem{J_TAC_tatarenko2018learning}
T.~Tatarenko, M.~Kamgarpour,
  \href{https://doi.org/10.1109/tac.2018.2841319}{Learning generalized Nash
  equilibria in a class of convex games}, IEEE Trans. Autom. Control 64~(4)
  (2018) 1426--1439.

\bibitem{C_NeurIPS_lanctot2017unified}
M.~Lanctot, V.~Zambaldi, et~al.,
  \href{https://proceedings.neurips.cc/paper/2017/hash/3323fe11e9595c09af38fe67567a9394-Abstract.html}{A
  unified game-theoretic approach to multiagent reinforcement learning}, in:
  NeurIPS, 2017.

\bibitem{B_hespanha2017noncooperative}
J.~P. Hespanha,
  \href{https://press.princeton.edu/books/hardcover/9780691175218/noncooperative-game-theory}{Noncooperative
  Game Theory: An Introduction for Engineers and Computer Scientists},
  Princeton University Press, 2017.

\bibitem{C_NIPS_goodfellow2014generative}
I.~Goodfellow, J.~{Pouget-Abadie}, et~al.,
  \href{https://proceedings.neurips.cc/paper/2014/hash/5ca3e9b122f61f8f06494c97b1afccf3-Abstract.html}{Generative
  adversarial nets}, in: NeurIPS, 2014.

\bibitem{J_NN_schmidhuber2020generative}
J.~Schmidhuber, \href{https://doi.org/10.1016/j.neunet.2020.04.008}{Generative
  adversarial networks are special cases of artificial curiosity (1990) and
  also closely related to predictability minimization (1991)}, Neural Netw. 127
  (2020) 58--66.

\bibitem{J_NRL_hildreth1957quadratic}
C.~Hildreth, \href{https://doi.org/10.1002/nav.3800040113}{A quadratic
  programming procedure}, Nav. Res. Log. Quarterly 4~(1) (1957) 79--85.

\bibitem{J_NRL_desopo1959convex}
D.~A. D'esopo, \href{https://doi.org/10.1002/nav.3800060105}{A convex
  programming procedure}, Nav. Res. Log. Quarterly 6~(1) (1959) 33--42.

\bibitem{J_MP_powell1973search}
M.~J.~D. Powell, \href{https://doi.org/10.1007/BF01584660}{On search directions
  for minimization algorithms}, Math. Program. 4~(1) (1973) 193--201.

\bibitem{J_AIJ_whitley1996evaluating}
D.~Whitley, S.~Rana, et~al.,
  \href{https://doi.org/10.1016/0004-3702(95)00124-7}{Evaluating evolutionary
  algorithms}, Artif. Intell. 85~(1) (1996) 245--276.

\bibitem{J_TEVC_wolpert1997no}
D.~Wolpert, W.~Macready, \href{https://doi.org/10.1109/4235.585893}{No free
  lunch theorems for optimization}, IEEE Trans. Evol. Comput. 1~(1) (1997)
  67--82.

\bibitem{B_leon2010linear}
S.~J. Leon, Linear Algebra with Applications, 8th Edition, Pearson, 2010.

\bibitem{B_boyd2004convex}
S.~P. Boyd, L.~Vandenberghe,
  \href{https://web.stanford.edu/~boyd/cvxbook/}{Convex Optimization},
  Cambridge University Press, 2004.

\bibitem{J_CACM_zaharia2016apache}
M.~Zaharia, R.~S. Xin, et~al., \href{https://doi.org/10.1145/2934664}{Apache
  spark: a unified engine for big data processing. a unified engine for big
  data processing}, Commun. ACM 59~(11) (2016) 56--65.

\bibitem{C_OSDI_moritz2018ray}
P.~Moritz, R.~Nishihara, et~al.,
  \href{https://www.usenix.org/conference/osdi18/presentation/moritz}{Ray: A
  distributed framework for emerging AI applications}, in: OSDI, 2018, pp.
  561--577.

\bibitem{J_JACM_holland1962outline}
J.~H. Holland, \href{https://doi.org/10.1145/321127.321128}{Outline for a
  logical theory of adaptive systems}, J. ACM 9~(3) (1962) 297--314.

\bibitem{BS_fogel1999overview}
D.~B. Fogel, \href{https://doi.org/10.1007/978-1-4612-1542-4_5}{An overview of
  evolutionary programming}, in: Evolutionary Algorithms, Springer, 1999, pp.
  89--109.

\bibitem{C_ICNN_kennedy1995particle}
J.~Kennedy, R.~Eberhart,
  \href{https://doi.org/10.1109/ICNN.1995.488968}{Particle swarm optimization},
  in: ICNN, 1995, pp. 1942--1948 vol.4.

\bibitem{J_JGO_storn1997differential}
R.~Storn, K.~Price, \href{https://doi.org/10.1023/A:1008202821328}{Differential
  evolution \textendash{} A simple and efficient heuristic for global
  optimization over continuous spaces}, J. Glob. Optim. 11~(4) (1997) 341--359.

\bibitem{B_larranaga2002estimation}
P.~Larra{\~n}aga (Ed.),
  \href{https://link.springer.com/book/10.1007/978-1-4615-1539-5}{Estimation of
  distribution algorithms: A new tool for evolutionary computation}, Springer,
  2002.

\bibitem{J_TEVC_loshchilov2019large}
I.~Loshchilov, T.~Glasmachers, H.~Beyer,
  \href{https://doi.org/10.1109/TEVC.2018.2855049}{Large scale black-box
  optimization by limited-memory matrix adaptation}, IEEE Trans. Evol. Comput.
  23~(2) (2019) 353--358.

\bibitem{C_Ecodyn_schwefel1988collective}
H.-P. Schwefel, \href{https://doi.org/10.1007/978-3-642-73953-8\_8}{Collective
  intelligence in evolving systems}, in: Ecodynamics, 1988, pp. 95--100.

\bibitem{C_PPSN_rudolph1990global}
G.~Rudolph, \href{https://doi.org/10.1007/BFb0029754}{Global optimization by
  means of distributed evolution strategies}, in: PPSN, 1990, pp. 209--213.

\bibitem{J_NatMI_kudela2022critical}
J.~Kudela, \href{https://doi.org/10.1038/s42256-022-00579-0}{A critical problem
  in benchmarking and analysis of evolutionary computation methods}, Nat. Mach.
  Intell. 4~(12) (2022) 1238--1245.

\bibitem{J_Nature_gould1993punctuated}
S.~J. Gould, N.~Eldredge, \href{https://doi.org/10.1038/366223a0}{Punctuated
  equilibrium comes of age}, Nature 366 (1993) 223--227.

\bibitem{C_GECCO_dejong2007introductory}
E.~D. De~Jong, K.~O. Stanley, R.~P. Wiegand,
  \href{https://doi.org/10.1145/1274000.1274108}{Introductory tutorial on
  coevolution}, in: GECCOC, 2007, pp. 3133--3157.

\bibitem{BS_anderson2008more}
P.~W. Anderson,
  \href{https://doi.org/10.7551/mitpress/9780262026215.003.0018}{More Is
  Different: Broken symmetry and the nature of the hierarchical structure of
  science.}, in: Emergence: Contemporary Readings in Philosophy and Science,
  The MIT Press, 2008, pp. 393--396.

\end{thebibliography}

\end{document}